%% file: bare_jrnl.tex
\documentclass[10pt,journal,compsoc]{IEEEtran}
\ifCLASSOPTIONcompsoc
  \usepackage[nocompress]{cite}
\else
  \usepackage{cite}
\fi
\ifCLASSINFOpdf
\else
\fi

\usepackage{hyperref}       
\usepackage{url}            
\usepackage{booktabs}       
\usepackage{amsfonts}       
\usepackage{nicefrac}       
\usepackage{microtype}      
\usepackage{xcolor}         

\input{math_commands.tex}

\usepackage{graphicx}
\usepackage{subcaption}
\usepackage{booktabs} 
\usepackage{enumitem}
\usepackage{xcolor}
\usepackage{array}

\newcommand\hll[1]{#1}
\newcommand\hlr[1]{#1}
\newcommand\hlrs[1]{#1}

\begin{document}
%
\title{\hll{Learning Interpretable Rules for Scalable \\Data Representation and Classification}}
%
%
%
%

\author{Zhuo~Wang,~
        Wei~Zhang,~\IEEEmembership{Member,~IEEE,}
        Ning~Liu,~
        and~Jianyong~Wang,~\IEEEmembership{Fellow,~IEEE}
\IEEEcompsocitemizethanks{\IEEEcompsocthanksitem Z. Wang and J. Wang are with the Department of Computer Science and Technology, Tsinghua University, Beijing 100084, China. Z. Wang is also with the Ant Group, Hangzhou 310058, China.\protect\\
E-mail: wang-z18@mails.tsinghua.edu.cn, jianyong@tsinghua.edu.cn.
\IEEEcompsocthanksitem W. Zhang is with the School of Computer Science and Technology, Shanghai Institute for AI Education, East China Normal University, Shanghai 200062, China. \protect\\
E-mail: zhangwei.thu2011@gmail.com
\IEEEcompsocthanksitem N. Liu is with the School of Software, Shandong University, Jinan 250101, China. 
\protect\\E-mail: victorliucs@gmail.com
\IEEEcompsocthanksitem Corresponding authors: W. Zhang 
 and J. Wang}
}

%
%


\markboth{IEEE TRANSACTIONS ON PATTERN ANALYSIS AND MACHINE INTELLIGENCE,~Vol.~46, No.~2, FEBRUARY~2024}%
{Wang \MakeLowercase{\textit{et al.}}: Learning Interpretable Rules for Scalable Data Representation and Classification}
%



\IEEEtitleabstractindextext{%
\begin{abstract}
Rule-based models, e.g., decision trees, are widely used in scenarios demanding high model interpretability for their transparent inner structures and good model expressivity. However, rule-based models are hard to optimize, especially on large data sets, due to their discrete parameters and structures. Ensemble methods and fuzzy/soft rules are commonly used to improve performance, but they sacrifice the model interpretability. To obtain both good scalability and interpretability, we propose a new classifier, named Rule-based Representation Learner (RRL), that automatically learns interpretable non-fuzzy rules for data representation and classification. To train the non-differentiable RRL effectively, we project it to a continuous space and propose a novel training method, called Gradient Grafting, that can directly optimize the discrete model using gradient descent. A \hll{novel} design of logical activation functions is also devised to increase the scalability of RRL and enable it to discretize the continuous features end-to-end. Exhaustive experiments on \hll{ten} small and four large data sets show that RRL outperforms the competitive interpretable approaches and can be easily adjusted to obtain a trade-off between classification accuracy and model complexity for different scenarios. 
 Our code is available at: \url{https://github.com/12wang3/rrl}.
\end{abstract}

}

\maketitle

\IEEEdisplaynontitleabstractindextext

%
\IEEEpeerreviewmaketitle

\IEEEraisesectionheading{\section{Introduction}\label{sec:introduction}}

\IEEEPARstart{A}{lthough} Deep Neural Networks (DNNs) have achieved impressive results in various machine learning tasks \cite{goodfellow2016deep}, rule-based models, benefiting from their transparent inner structures and good model expressivity, still play an important role in domains demanding high model interpretability, such as medicine, finance, and politics \cite{doshi2017towards}.
In practice, rule-based models can easily provide explanations for users to earn their trust and help protect their rights \cite{molnar2019,lipton2016mythos}. By analyzing the learned rules, practitioners can understand the decision mechanism of models and use their knowledge to improve or debug the models \cite{chu2018exact}.
Moreover, even if post-hoc methods can provide interpretations for DNNs, the interpretations from rule-based models are more faithful and specific \cite{murdoch2019interpretable}.
However, conventional rule-based models are hard to optimize, especially on large data sets, due to their discrete parameters and structures, which limit their application scope. 
To take advantage of rule-based models in more scenarios, we urgently need to answer such a question: \textit{how to improve the scalability of rule-based models while keeping their interpretability?}

Studies in recent years provide some solutions to improve conventional rule-based models in different aspects. Ensemble methods and soft/fuzzy rules are proposed to improve the performance and scalability of rule-based models but at the cost of model interpretability \cite{ke2017lightgbm,breiman2001random,irsoy2012soft}. Bayesian frameworks are also leveraged to more reasonably restrict and adjust the structures of rule-based models \cite{letham2015interpretable,wang2017bayesian,yang2017scalable}. However, due to the non-differentiable model structure, they have to use methods like MCMC or Simulated Annealing, which could be time-consuming for large models. 
Another way to improve rule-based models is to let a high-performance but complex model (e.g., DNN) teach a rule-based model \cite{frosst2017distilling,ribeiro2016should}. However, learning from a complex model requires soft rules, or the fidelity of the student model is not guaranteed. 
The recent study~\cite{wang2020transparent} tries to extract hierarchical rule sets from a tailored neural network. 
When the network is large, the extracted rules could behave quite differently from the neural network and become useless in most cases. 
Nevertheless, combined with binarized networks \cite{cour2015bconnect}, it inspires us that we can search for the discrete solutions of interpretable rule-based models in a continuous space leveraging effective optimization methods like gradient descent.

In this paper, we propose a novel rule-based model named \textbf{Rule-based Representation Learner (RRL)} (see Figure~\ref{fig:RRL}). We summarize the key contributions as follows:

\begin{itemize}[leftmargin=*,itemsep=0pt,parsep=0.0em,topsep=0.0em,partopsep=0.0em]
\item To achieve good model \textbf{transparency and expressivity}, RRL is formulated as a hierarchical model, with layers supporting automatic feature discretization, rule-based representation learning in flexible conjunctive and disjunctive normal forms, and rule importance evaluation.

\item To facilitate \textbf{training effectiveness}, RRL exploits a novel gradient-based discrete model training method, Gradient Grafting, that directly optimizes the discrete model and uses the gradient information at both continuous and discrete parametric points to accommodate more scenarios. 
\hll{For deep RRL, Hierarchical Grafting are also proposed.}

\item To ensure \textbf{data scalability}, \hll{we propose novel logical activation functions for RRL to handle high-dimensional features with fewer resources and higher speed.} By further combining the novel logical activation functions with a tailored feature binarization layer, it realizes the continuous feature discretization end-to-end.

\item We conduct experiments on \hll{ten} small data sets and four large data sets to validate the advantages, i.e., good \textbf{accuracy and interpretability}, of our model over other representative classification models. The benefits of the model's key components are verified by the experiments.
\end{itemize}

\section{Related Work}
\textbf{Rule-based Models}. Decision tree, rule list, and rule set are the widely used structures in rule-based models. For their discrete parameters and non-differentiable structures, we have to train them by employing various heuristic methods \cite{Quinlan:1993:CPM:583200,breiman2017classification,cohen1995fast,wei2019generalized}, which may not find the globally best solution or a solution with close performance. Alternatively, train them with search algorithms \cite{wang2017bayesian,angelino2017learning,lin2020generalized}, which could take too much time on large data sets. 
In recent studies, Bayesian frameworks are leveraged to restrict and adjust model structure more reasonably \cite{letham2015interpretable,wang2017bayesian,yang2017scalable}.
\cite{lakkaraju2016interpretable} learns independent if-then rules with smooth local search. 
However, except for heuristic methods, most existing rule-based models need frequent itemsets mining and/or long-time searching, which limits their applications. Moreover, it is hard for these rule-based models to get comparable performance with complex models like Random Forest.

Ensemble models like Random Forest \cite{breiman2001random} and Gradient Boosted Decision Trees \cite{chen2016xgboost,ke2017lightgbm} have better performance than the single rule-based model. However, since the decision is made by hundreds of models, ensemble models are commonly not considered as interpretable models \cite{hara2016making}. 
\hlr{Another way to improve rule-based model performance is changing the form of rules, e.g., soft or fuzzy rules \cite{irsoy2012soft,ishibuchi2005rule}. Soft rules, e.g., soft decision trees \cite{irsoy2012soft}, use logistic regressions to make soft decisions at each step. Fuzzy rules use fuzzy sets to build each rule \cite{ishibuchi2005rule}. However, soft or fuzzy rules also sacrifice interpretability since non-discrete rules are much harder to understand than discrete ones.}
Deep Neural Decision Tree \cite{yang2018deep} is a tree model realized by neural networks with the help of soft binning function and Kronecker product. However, due to the use of Kronecker product, it is not scalable with respect to the number of features.
\hll{Neuro-symbolic models that focus on Inductive Logic Programming (ILP) problems are also proposed recently\hlr{~\cite{glanois2022neuro,cheng2022rlogic,zimmer2021differentiable,chaudhury-etal-2023-learning}}. However, most of them heavily depend on hand-crafted formulas or Knowledge Graphs.
DeepLogic \cite{duan2022deeplogic} and \hlr{Abductive Learning \cite{zhou2021abductive,dai2019bridging}} jointly learn neural perception and logical reasoning through mutual supervision signals. However, the tasks that they can handle are limited due to their structure, and the neural perception part is a black box that we can hardly understand.}
Other studies try to teach the rule-based model by a complex model, e.g., DNN, or extract rule-based models from complex models \cite{frosst2017distilling,ribeiro2016should,zhang2020mining,wang2020transparent}. 
\hlr{For example, FIRE~\cite{Liu23fire} utilizes an optimization framework to select sparse representative subsets of rules from ensemble tree models like random forest to enhance interpretability.}
However, the performance of the student models or extracted models is significantly inferior to complex models. 

\textbf{Gradient-based Discrete Model Training}. The gradient-based discrete model training methods are mainly proposed to train binary or quantized neural networks for network compression and acceleration.
\cite{cour2015bconnect,courbariaux2016binarized} propose to use the Straight-Through Estimator (STE) for binary neural network training. However, STE requires gradient information at discrete points, which limits its applications. 
ProxQuant \cite{bai2018proxquant} formulates quantized network training as a regularized learning problem and optimizes it via the \hlr{prox-gradient \cite{bai2018proxquant}} method. ProxQuant can use gradients at non-discrete points but cannot directly optimize for the discrete model.
The \hlr{Random Binarization (RB)} method \cite{wang2020transparent} trains a neural network with random binarization for its weights to ensure the discrete and the continuous model behave similarly.
However, when the model is large, the differences between the discrete and the continuous model are inevitable. Gumbel-Softmax estimator \cite{jang2016categorical} generates a categorical distribution with a differentiable sample. 
However, it can hardly deal with a large number of variables, e.g., the weights of binary networks, for it is a biased estimator. 
Our method, i.e., Gradient Grafting, is different from all the aforementioned works in using gradient information of both discrete and continuous models in each backpropagation.
\section{Rule-based Representation Learner}
\noindent \textbf{Notation Description.}
Let $\sD=\{(X_1,Y_1),\dots,(X_N,Y_N)\}$ denote the training data set with $N$ instances, where $X_i$ is the observed feature vector of the $i$-th instance with the $j$-th entry as $X_{i,j}$, and $Y_i$ is the corresponding one-hot class label vector, $i\in\{1, \dots, N\}$.
Each feature value can be either discrete or continuous.
All the classes take discrete values, and the number of class labels is denoted by $M$.
We use one-hot encoding to represent all discrete features as binary features.
Let $C_i\in \R^m$ and $B_i\in \{0,1\}^b$ denote the continuous feature vector and the binary feature vector of the $i$-th instance respectively.
Therefore, $X_i=C_i\oplus B_i$, where $\oplus$ represents the operator that concatenates two vectors.
Throughout this paper, we use 1 (True) and 0 (False) to represent the two states of a Boolean variable. Thus each dimension of a binary feature vector corresponds to a Boolean variable.
\begin{figure}
    \centering
    \includegraphics[width=0.49\textwidth]{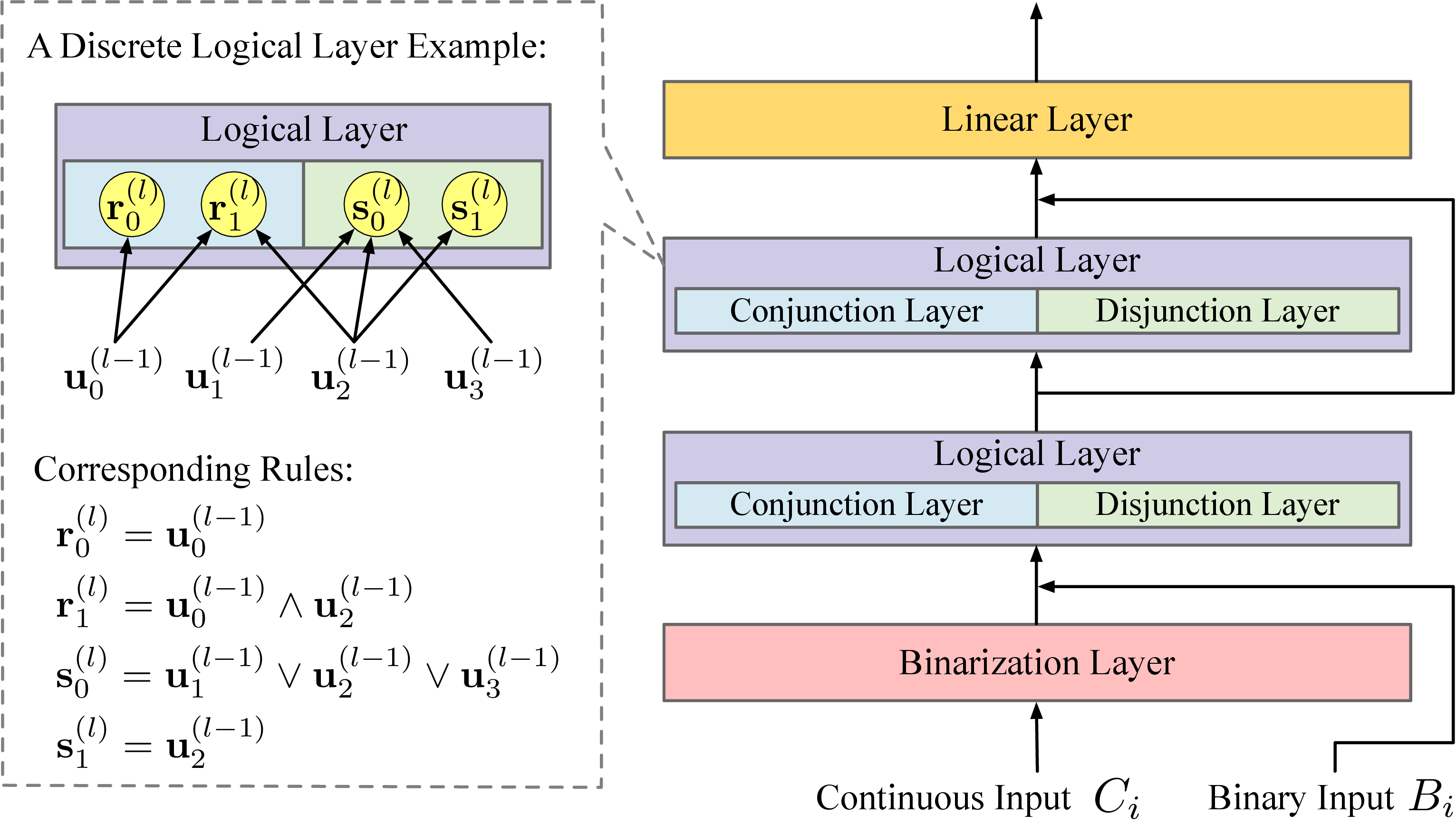}
    \caption{A Rule-based Representation Learner example. The dashed box shows an example of a discrete logical layer and its corresponding rules.}
    \label{fig:RRL}
\end{figure}

\noindent \textbf{Overall Structure.}
A Rule-based Representation Learner (RRL), denoted by $\mathcal{F}$, is a hierarchical model consisting of three different types of layers. Each layer in RRL not only contains a specific number of nodes but also has trainable edges connected with its previous layer. Let $\mathcal{U}^{(l)}$ denote the $l$-th layer of RRL, $\mathbf{u}^{(l)}_{j}$ indicate the $j$-th node in the layer, and $\vn_l$ represent the corresponding number of nodes, $l\in \{0,\dots,L\}$. The output of the $l$-th layer is a vector containing the values of all the nodes in the layer.
For ease of expression, we denote this vector by $\mathbf{u}^{(l)}$.
There are only one binarization layer, i.e., $\mathcal{U}^{(0)}$,  and one linear layer, i.e., $\mathcal{U}^{(L)}$, in RRL, but the number of middle layers, i.e., logical layers, can be flexibly adjusted according to the specific situation. The logical layers mainly aim to learn the non-linear part of the data, while the linear layer aims to learn the linear part. One example of RRL is shown in Figure \ref{fig:RRL}.

When we input the $i$-th instance to RRL, the binarization layer will first binarize the continuous feature vector $C_i$ into a new binary vector $\bar{C}_i$. Then, $\bar{C}_i$ and $B_i$ are concatenated together as $\mathbf{u}^{(0)}$ and inputted to the first logical layer. 
The logical layers are designed to automatically learn data representations using logical rules, and the stacked logical layers can learn rules in more complex forms. After going through all the logical layers, the output of the last logical layer can be considered as the new feature vector to represent the instance, wherein each feature corresponds to one rule formulated by the original features.
As such, the whole RRL is composed of a feature learner and a linear classifier (linear layer). 
Moreover, the skip connections in RRL can skip unnecessary logical layers.
\hlr{
For example, if one RRL has three logical layers (i.e., $\mathcal{U}^{(1)}$, $\mathcal{U}^{(2)}$, and $\mathcal{U}^{(3)}$) and a skip connection used to skip $\mathcal{U}^{(2)}$, then the output of $\mathcal{U}^{(1)}$ can be directly achieved by $\mathcal{U}^{(3)}$. In other words, the output vector of $\mathcal{U}^{(1)}$ and $\mathcal{U}^{(2)}$ are concatenated together as the input of $\mathcal{U}^{(3)}$. Therefore, if $\mathcal{U}^{(2)}$ is unnecessary, $\mathcal{U}^{(3)}$ can only use the output of $\mathcal{U}^{(1)}$.
}
In what follows, the details of these components will be elaborated.
\subsection{Logical Layer}
\label{section:logical_layer}
Considering the binarization layer needs the help of its following logical layer to binarize features in an end-to-end way, we introduce logical layers first. 
As mentioned above, logical layers can learn data representations using logical rules automatically. To achieve this, logical layers are designed to have a discrete version and a continuous version. The discrete version is used for training, testing and interpretation while the continuous version is only used for training. 
It is worth noting that the discrete RRL indicates the parameter weights of logical layers take discrete values (i.e., 0 or 1) while the parameter weights and biases of the linear layer still take continuous values.

\subsubsection{Discrete Version}
One logical layer consists of one conjunction layer and one disjunction layer. In discrete version, let $\mathcal{R}^{(l)}$ and $\mathcal{S}^{(l)}$ denote the conjunction and disjunction layer of $\mathcal{U}^{(l)}$ ($l\in \{1,2,\dots, L-1\}$) respectively. We denote the $i$-th node in $\mathcal{R}^{(l)}$ by $\mathbf{r}_{i}^{(l)}$, and the $i$-th node in $\mathcal{S}^{(l)}$ by $\mathbf{s}_{i}^{(l)}$.
Specifically speaking, node $\mathbf{r}_{i}^{(l)}$ corresponds to the conjunction of nodes in the previous layer connected with $\mathbf{r}_{i}^{(l)}$, 
while node $\mathbf{s}_{i}^{(l)}$ corresponds to the disjunction of nodes in previous layer connected with $\mathbf{s}_{i}^{(l)}$.
Formally, the two types of nodes are defined as follows:
\begin{equation}
\label{eq:discrete_logic}
\mathbf{r}_{i}^{(l)} =
\bigwedge_{W_{i,j}^{(l,0)}=1}\mathbf{u}_{j}^{(l-1)},\;\;\;\;\;\; \mathbf{s}_{i}^{(l)} = \bigvee_{W_{i,j}^{(l,1)}=1}\mathbf{u}_{j}^{(l-1)},
\end{equation}
where $W^{(l,0)}$ denotes the adjacency matrix of the conjunction layer $\mathcal{R}^{(l)}$ and the previous layer $\mathcal{U}^{(l-1)}$, and $W_{i,j}^{(l,0)} \in \{0,1\}$. $W_{i,j}^{(l,0)}=1$ indicates there exists an edge connecting $\mathbf{r}_{i}^{(l)}$ to $\mathbf{u}_{j}^{(l-1)}$, otherwise $W_{i,j}^{(l,0)}=0$. Similarly, $W^{(l,1)}$ is the adjacency matrix of the disjunction layer $\mathcal{S}^{(l)}$ and $\mathcal{U}^{(l-1)}$. Similar to neural networks, we regard these adjacency matrices as the weight matrices of logical layers. $\mathbf{u}^{(l)}=\mathbf{r}^{(l)}\oplus \mathbf{s}^{(l)}$, where $\mathbf{r}^{(l)}$ and $\mathbf{s}^{(l)}$ are the outputs of $\mathcal{R}^{(l)}$ and $\mathcal{S}^{(l)}$ respectively.

The function of the logical layer is similar to the notion ``level'' in \cite{wang2020transparent}. However, one level in that work, which actually consists of two layers, can only represent rules in Disjunctive Normal Form (DNF), while two logical layers in RRL can represent rules in DNF and Conjunctive Normal Form (CNF) at the same time. Connecting nodes in $\mathcal{R}^{(l)}$ with nodes in $\mathcal{S}^{(l-1)}$, we get rules in CNF, while connecting nodes in $\mathcal{S}^{(l)}$ with nodes in $\mathcal{R}^{(l-1)}$, we get rules in DNF. The flexibility of logical layer is quite important. For instance, the length of CNF rule $(\va_1\vee \va_2)\wedge \dots \wedge(\va_{2n-1}\vee \va_{2n})$ is $2n$, but the length of its corresponding DNF rule $(\va_1\wedge \va_3 \dots \wedge \va_{2n-1})\vee \dots \vee(\va_2\wedge \va_4 \dots \wedge \va_{2n})$ is $n\cdot2^{n}$, which means layers that only represent DNF can hardly learn this CNF rule.

\subsubsection{Continuous Version}
Although the discrete logical layers have good interpretability, they are hard to train for their discrete parameters and non-differentiable structures. 
Inspired by the training process of binary neural networks that searches the discrete solution in a continuous space\hlr{\cite{cour2015bconnect,courbariaux2016binarized}}, we extend the discrete logical layer to a continuous version. The continuous version is differentiable, and when we discretize the parameters of a continuous logical layer, we can obtain its corresponding discrete logical layer.
Therefore, the continuous logical layer and its corresponding discrete logical layer can also be considered as sharing the same parameters, but the discrete logical layer needs to discretize the parameters first.

Let $\hat{\mathcal{U}}^{(l)}$ denote the continuous logical layer, and $\hat{\mathcal{R}}^{(l)}$  and $\hat{\mathcal{S}}^{(l)}$ denote the continuous conjunction and disjunction layer respectively, $l\in \{1,2,\dots, L-1\}$. Let $\hat{W}^{(l,0)}$ and $\hat{W}^{(l,1)}$ denote the weight matrices of $\hat{\mathcal{R}}^{(l)}$  and $\hat{\mathcal{S}}^{(l)}$ respectively. $\hat{W}_{i,j}^{(l,0)},\hat{W}_{i,j}^{(l,1)}\in [0,1]$. To make the whole Equation \ref{eq:discrete_logic} differentiable, we leverage the Logical Activation Functions (LAF) proposed by \cite{payani2019learning}:
\begin{equation}
\label{eq:conj_disj}
\begin{aligned}
\textit{Conj}(\mathbf{h},
W_{i})&=\prod_{j=1}^{n}F_{c}(\mathbf{h}_{j}, W_{i,j}), \\
\textit{Disj}(\mathbf{h}, W_{i})&=1-\prod_{j=1}^{n}(1-F_{d}(\mathbf{h}_{j}, W_{i,j})),
\end{aligned}
\end{equation}
where $F_{c}(h,w)=1-w(1-h)$ and $F_{d}(h,w)=h\cdot w$. In Equation \ref{eq:conj_disj}, if  $\mathbf{h}$ and $W_{i}$ are both binary vectors, then $\textit{Conj}(\mathbf{h}, W_{i})=\bigwedge_{W_{i,j}=1}\mathbf{h}_{j}$ and $\textit{Disj}(\mathbf{h}, W_{i})=\bigvee_{W_{i,j}=1}\mathbf{h}_{j}$. $F_c(h,w)$ and $F_d(h,w)$ decide how much $\mathbf{h}_j$ would affect the operation according to $W_{i,j}$. If $W_{i,j}=0$, $\mathbf{h}_j$ would have no effect on the operation.
Equation \ref{eq:conj_disj} can be considered as extensions to t-norm fuzzy logic \cite{hajek2013metamathematics}.
After using continuous weights and logical activation functions, the nodes in $\hat{\mathcal{R}}^{(l)}$ and $\hat{\mathcal{S}}^{(l)}$ are defined as follows:
\begin{equation}
\label{eq:conj_hat_disj_hat}
\hat{\mathbf{r}}_{i}^{(l)} =\textit{Conj}(\hat{\mathbf{u}}^{(l-1)}, \hat{W}_{i}^{(l,0)}), \;\;\;
\hat{\mathbf{s}}_{i}^{(l)}=\textit{Disj}(\hat{\mathbf{u}}^{(l-1)}, \hat{W}_{i}^{(l,1)})
\end{equation}

\hll{
For ease of expression, let $\textit{LAF}(\cdot)$ denote the combination of all the $\textit{Conj}(\cdot)$ and $\textit{Disj}(\cdot)$ functions in one continuous logical layer. Let $\hat{W}^{(l)}$ denote the concatenation of $\hat{W}^{(l,0)}$ and $\hat{W}^{(l,1)}$, i.e., $\hat{W}^{(l)}=[ \hat{W}^{(l,0)},\hat{W}^{(l,1)} ]$. According to Equation \ref{eq:conj_hat_disj_hat} and the equation $\hat{\mathbf{u}}^{(l)}=\hat{\mathbf{r}}^{(l)}\oplus \hat{\mathbf{s}}^{(l)}$, we can compute $\hat{\mathbf{u}}^{(l)}$ using $\hat{\mathbf{u}}^{(l-1)}$ by:
\begin{equation}
\label{eq:laf_q}
\hat{\mathbf{u}}^{(l)} =\textit{LAF}(\hat{\mathbf{u}}^{(l-1)}, \hat{W}^{(l)})
\end{equation}
}
Now the whole logical layer is differentiable and can be trained by gradient descent. However, the above logical activation functions suffer from the serious vanishing gradient problem \hlr{\cite{Learning279181}}. The main reason can be found by analyzing the partial derivative of each node w.r.t. its directly connected weights and w.r.t. its directly connected nodes as follows:
\begin{equation}
\label{eq:derivative_rw_ru}
\begin{aligned}
\frac{\partial \hat{\mathbf{r}}_{i}^{(l)}}{\partial \hat{W}_{i,j}^{(l,0)}}
&=(\hat{\mathbf{u}}_{j}^{(l-1)}-1) \cdot \prod_{k \neq j}F_c(\hat{\mathbf{u}}_{k}^{(l-1)}, \hat{W}_{i,k}^{(l,0)}),  \\
\frac{\partial \hat{\mathbf{r}}_{i}^{(l)}}{\partial \hat{\mathbf{u}}_{j}^{(l-1)}}
&=\hat{W}_{i,j}^{(l,0)} \cdot \prod_{k \neq j}F_c(\hat{\mathbf{u}}_{k}^{(l-1)}, \hat{W}_{i,k}^{(l,0)})
\end{aligned}
\end{equation}
Since $\hat{\mathbf{u}}_{k}^{(l-1)}$ and $\hat{W}_{i,k}^{(l,0)}$ are in the range $[0, 1]$, the values of $F_c(\cdot)$ in Equation \ref{eq:derivative_rw_ru} are in the range $[0, 1]$ as well. If $\mathbf{n}_{l-1}$ is large and most of the values of $F_c(\cdot)$ are not 1, then the derivative is close to 0 due to the multiplications (See Appendix \ref{vanishing_gradient_problem} for the analysis of $\hat{\mathbf{s}}_{i}^{(l)}$). \cite{wang2020transparent} tries to use weight initialization to make $F_c(\cdot)$ close to 1 at the beginning. However, when dealing with hundreds of features, the vanishing gradient problem is still inevitable.

\hll{\subsubsection{Novel Logical Activation Functions}
\label{sec:NLAF}
\noindent \textbf{Vanishing Gradient Prevention.}}
We found that using the multiplications to simulate the logical operations in Equation \ref{eq:conj_disj} is the main reason for vanishing gradients and propose an improved design of logical activation functions. One straightforward idea is to convert multiplications into additions using logarithm, e.g., $\log(\prod_{j=1}^{n}F_{c}(\mathbf{h}_{j}, W_{i,j}))=\sum_{j=1}^{n}\log(F_{c}(\mathbf{h}_{j}, W_{i,j}))$. However, after taking the logarithm, the logical activation functions in Equation \ref{eq:conj_disj} cannot keep the characteristics of logical operations any more, and the ranges of $Conj(\cdot)$ and $Disj(\cdot)$ are not $[0,1]$. To deal with this problem, we need a projection function to fix it. Apparently, the inverse function of $\log(x)$, i.e., $e^x$, is not suitable.

For the projection function $g$, three conditions must be satisfied: (\romannumeral1) $g(0)=e^0.$ (\romannumeral2) $\lim_{x\rightarrow -\infty}g(x)=\lim_{x\rightarrow -\infty }e^x=0.$ (\romannumeral3) $\lim_{x\rightarrow -\infty}\frac{e^x}{g(x)}=0.$ Condition (\romannumeral1) and (\romannumeral2) aim to keep the range and tendency of logical activation functions. Condition (\romannumeral3) aims to lower the speed of approaching zero when $x\rightarrow -\infty$.
In this work, we choose $g(x)=\frac{1}{1-x}$ as the projection function, and the improvement of logical activation functions can be summarized as the function $\mathbb{P}(v)=\frac{1}{1-\log(v)}$. The improved conjunction function $\textit{Conj}_{+}$ and disjunction function $\textit{Disj}_{+}$ are given by:
\begin{equation}
\label{eq:conj+_disj+}
\begin{aligned}
\textit{Conj}_{+}(\mathbf{h},
W_{i})&= \mathbb{P}(\prod_{j=1}^{n}(F_{c}(\mathbf{h}_{j}, W_{i,j})+\epsilon)), \\ \textit{Disj}_{+}(\mathbf{h}, W_{i})&=1-\mathbb{P}(\prod_{j=1}^{n}(1-F_{d}(\mathbf{h}_{j}, W_{i,j})+\epsilon)),
\end{aligned}
\end{equation}
where $\epsilon$ is a small constant, e.g., $10^{-10}$. The improved logical activation functions can avoid the vanishing gradient problem in most scenarios and are much more scalable than the originals.
Moreover, considering that $\frac{d \mathbb{P}(v)}{d v}=\frac{\mathbb{P}^2(v)}{v}$, when $n$ in Equation \ref{eq:conj+_disj+} is extremely large, $\frac{d \mathbb{P}(v)}{d v}$ may be very close to 0 due to $\mathbb{P}^2(v)$. One trick to deal with it is replacing $\frac{\mathbb{P}^2(v)}{v}$ with $\frac{\mathbb{P}(\mathbb{P}^2(v))}{v}$ for $\mathbb{P}$ can lower the speed of approaching 0 while keeping the value range and tendency.

\begin{table}[!t]
  \caption{\hll{A generalized truth table. $\infty$ represents a sufficiently large value, rather than just the infinity, for ease of expression.}}
  \label{tab:truth_table}
 \scalebox{0.9}{
 \hll{
  \begin{tabular}{cc|cccc}
\toprule
$h$ & $w$ & $1-hw$	& $\log(1-hw)$ & $-\mathcal{G}(h)\cdot \mathcal{G}(w)$ & $\mathcal{P}(-\mathcal{G}(h)\cdot \mathcal{G}(w))$\\
\midrule									
0	&	0	&	1	&	0	&	$-0\cdot 0=0$	& 1\\
0	&	1	&	1	&	0	&	$-0\cdot (-\infty)= 0$	& 1\\
1	&	0	&	1	&	0	&	$+\infty \cdot 0 = 0$	& 1\\
1	&	1	&	0	&	$-\infty$	&	$+\infty \cdot (-\infty) \approx -\infty$	& 0\\
\bottomrule
\end{tabular}
}
}
\end{table}
\hll{
\noindent \textbf{Coupled Computation.}
Although Equation \ref{eq:conj+_disj+} could overcome the vanishing gradient problem, its scalability is still limited since its computation is coupled. This coupling is inherited from Equation \ref{eq:conj_disj} for both Equation \ref{eq:conj_disj} and \ref{eq:conj+_disj+} are based on the following basic computation:
\begin{equation}
\label{eq:basic_laf_eq}
\prod_{j=1}^{n}(1-\mathbf{h}_{j}\cdot W_{i,j}) \;\;\;\text{or}\;\;\; \sum_{j=1}^{n}\log(1-\mathbf{h}_{j}\cdot W_{i,j})
\end{equation}
Equation \ref{eq:basic_laf_eq} is coupled since $(1-\mathbf{h}_{j}\cdot W_{i,j})$ cannot be factored into a product of two terms that only contains one variable (i.e., $\mathbf{h}_{j}$ or $W_{i,j}$) in each term. 
Therefore, in practice, we cannot implement this equation by effective operations like matrix multiplication when using deep learning frameworks, e.g., TensorFlow \cite{abadi2016tensorflow} and PyTorch \cite{paszke2019pytorch}, especially for mini-batch training. 
One way to implement Equation \ref{eq:basic_laf_eq} is using the broadcasting. However, for a mini-batch matrix $H_{m \times n}$ and a weight matrix $W_{l \times n}$, the broadcasting would generate an intermediate result matrix $A_{m \times n \times l}$, which could cost lots of GPU memory and slow down the whole computation.
Another way is designing tailored CUDA kernels \cite{cook2012cuda} that avoid the intermediate result matrix. However, CUDA kernels with effective data parallelism are hard to design, and we cannot benefit from the existing optimization for matrix multiplication in this way.
}

\hll{
\noindent \textbf{Computation Decoupling.} To tackle the aforementioned challenges and improve the scalability of our model, we propose novel logical activation functions that can be implemented by matrix multiplications.
}

\hll{
Inspired by Equation \ref{eq:conj_disj}, we find the key point of the logical activation function is the design of the short circuit and open circuit. The short circuit indicates that we can exclude the variables from the logical operations according to their directly connected weights, i.e., having no effect on the final results if the weight is zero. The open circuit indicates that all the included variables can directly determine the final results. For example, a 0 (False) is an open circuit for the conjunction, while a 1 (True) is an open circuit for the disjunction.
The first three columns of Table \ref{tab:truth_table} show the truth table of the multiplication term in Equation \ref{eq:basic_laf_eq}, i.e., $1-hw$.
We can see that the basic computation of Equation \ref{eq:conj_disj}, i.e., Equation \ref{eq:basic_laf_eq}, simulates the open circuit by multiplying zero and simulates the short circuit by replacing the excluded variables with one.
}

\hll{
To obtain the same effect using matrix multiplication, we can simulate the open circuit by adding the negative infinity and simulate the short circuit by replacing the excluded variables with zero. Specifically, we adopt a function $\mathcal{G}(\cdot)$ to simulate the open circuit and the short circuit and update the aforementioned projection function $g(\cdot)$ as $\mathcal{P}(\cdot)$ to keep the characteristics of logical operations. These two functions are defined as follows:
\begin{equation}
\mathcal{G}(x)=1-\frac{1}{1-(\alpha x)^{\beta}},\;\;\;\;\;\;
\mathcal{P}(x)=\frac{1}{(1-x)^\gamma},
\end{equation}
where $\alpha$, $\beta$, and $\gamma$ are all constants. $\alpha$ is used to change the value range and avoid the division by zero error, while $\beta$ is used to change the trend of the values. We set $\alpha$ close to one and set $\beta$ according to $\alpha$, e.g., $\alpha=0.999, \beta=8$. $\gamma$ aims to control the speed of the projection function approaching zero when $x\rightarrow -\infty$. When $\gamma=1$, $\mathcal{P}(\cdot)$ equals $g(\cdot)$. 
In practice, we find $(\alpha, \beta, \gamma)\in \{(0.999, 8, 1), (0.999, 8, 3), (0.9, 3, 3)\}$ is applicable for most scenarios \hlr{(see Appendix \ref{appendix:alpha_beta_gamma} for sensitivity studies of $\alpha$, $\beta$, $\gamma$)}.
With $\mathcal{G}(\cdot)$ and $\mathcal{P}(\cdot)$, the basic computation for the novel logical activation functions is given by:
\begin{equation}
\label{eq:basic_laf_mm}
\mathcal{Q}(\mathbf{h}, W_{i})=\mathcal{P}\Bigl (-\sum_{j=1}^{n}\mathcal{G}(\mathbf{h}_{j})\cdot \mathcal{G}(W_{i,j})\Bigr ) 
\end{equation}
Table \ref{tab:truth_table} also shows a generalized truth table of the term in Equation \ref{eq:basic_laf_mm}, and we can leverage it to better understand Equation \ref{eq:basic_laf_mm}.
For ease of expression, let $\infty$ denote a sufficiently large value instead of only the infinity.
We can see that when $W_{i,j}=0$, the short circuit is simulated by multiplying $\mathcal{G}(\mathbf{h}_{j})$ by $\mathcal{G}(W_{i,j})=0$.
When $W_{i,j}=1$, if $\mathbf{h}_{j}=1$, it will simulate the open circuit by adding $-\mathcal{G}(\mathbf{h}_{j})\cdot \mathcal{G}(W_{i,j})=-\infty$.
When $\mathbf{h}_{j},W_{i,j}\in (0,1)$, $\mathcal{G}(W_{i,j})$ and $-\mathcal{G}(\mathbf{h}_{j})\cdot \mathcal{G}(W_{i,j})$ use a value in the range $(-\infty, 0)$ to measure the degree of the short circuit and open circuit, respectively.
$\mathcal{P}(\cdot)$ projects the value range from $(-\infty, 0]$ to $(0, 1]$.
}

\hll{
The advantage of Equation \ref{eq:basic_laf_mm} is that it can decouple the computation using $\mathcal{G}(\cdot)$. Now, the whole equation can be easily converted into a matrix multiplication form for the mini-batch training. Let $H_{m \times n}$ denote the mini-batch matrix and $W_{l \times n}$ denote the weight matrix. The matrix multiplication form is given by:
\begin{equation}
\mathcal{Q}(H, W)=\mathcal{P}\Bigl(-\mathcal{G}(H)\cdot \mathcal{G}(W^\top) \Bigr),
\end{equation}
where $\mathcal{G}(\cdot)$ and $\mathcal{P}(\cdot)$ are element-wise functions (e.g., if $A=\mathcal{G}(H)$, then $A_{i,j}=\mathcal{G}(H_{i,j})$).
With the novel basic computation, i.e., Equation \ref{eq:basic_laf_mm}, Equation \ref{eq:conj+_disj+} can be updated as follows:
\begin{equation}
\label{eq:mm_conj+_disj+}
\begin{aligned}
\textit{Conj}_{+}(\mathbf{h},
W_{i})&=\mathcal{Q}(1-\mathbf{h}, W_{i}), \\  \textit{Disj}_{+}(\mathbf{h}, W_{i})&=1-\mathcal{Q}(\mathbf{h}, W_{i})
\end{aligned}
\end{equation}
Compared with the original logical activation functions, the novel logical activation functions, i.e., Equation \ref{eq:mm_conj+_disj+}, not only overcome the vanishing gradient problem but also take less GPU memory and run faster. These advantages are verified in the experiments.
}

\subsection{Binarization Layer}
The binarization layer is mainly used to divide the continuous feature values into several bins. By combining one binarization layer and one logical layer, we can automatically choose the appropriate bins for feature discretization (binarization), i.e., binarizing features in an end-to-end way.

For the $j$-th continuous feature to be binarized, there are $k$ lower bounds ($\mathcal{T}_{j,1}, \dots, \mathcal{T}_{j,k}$) and $k$ upper bounds ($\mathcal{H}_{j,1}, \dots, \mathcal{H}_{j,k}$). All these bounds are randomly selected (e.g., from uniform distribution) in the value range of the $j$-th continuous feature, and these bounds are not trainable. When inputting one continuous feature vector $\vc$, the binarization layer will check if $\vc_{j}$ satisfies the bounds and get the following binary vector:
\begin{equation}
\resizebox{1.03\hsize}{!}{$Q_{j} = [q(\vc_{j}-\mathcal{T}_{j,1}),\dots, q(\vc_{j}-\mathcal{T}_{j,k}),q(\mathcal{H}_{j,1}-\vc_{j}),\dots, q(\mathcal{H}_{j,k}-\vc_{j})],$}
\end{equation}
where $q(x)= \1_\mathrm{x>0}$. \hlr{$Q_{j}$ is the result of comparing $\bm{c}_{j}$ with all the bounds. For example, $q(\bm{c}_{j}-\mathcal{T}_{j,1})=\bm{1}_\mathrm{\bm{c}_{j}-\mathcal{T}_{j,1}>0}=\bm{1}_\mathrm{\bm{c}_{j}>\mathcal{T}_{j,1}}$, i.e., $q(\bm{c}_{j}-\mathcal{T}_{j,1})$ represents if $\bm{c}_{j}$ is greater than the bound $\mathcal{T}_{j,1}$.}
If we input the $i$-th instance, i.e., $\vc=C_i$, then $\bar{C}_i=Q_1\oplus Q_2 \dots \oplus Q_{m}$ and $\mathbf{u}^{(0)}=\bar{C}_i\oplus B_i$.
After inputting $\mathbf{u}^{(0)}$ to the logical layer $\mathcal{U}^{(1)}$, the edge connections between $\mathcal{U}^{(1)}$ and $\mathcal{U}^{(0)}$ indicate the choice of bounds (bins). For example, if $\mathbf{r}_{i}^{(1)}$ is connected to the nodes corresponding to $q(\vc_{j}-\mathcal{T}_{j,1})$ and $q(\mathcal{H}_{j,2}-\vc_{j})$, then $\mathbf{r}_{i}^{(1)}$ contains the bin $(\mathcal{T}_{j,1}<\vc_{j})\wedge(\vc_{j}<\mathcal{H}_{j,2})$. If we replace $\mathbf{r}_{i}^{(1)}$ with $\mathbf{s}_{i}^{(1)}$ in the example, we can get $(\mathcal{T}_{j,1}<\vc_{j})\vee(\vc_{j}<\mathcal{H}_{j,2})$. It should be noted that, in practice, if $\mathcal{T}_{j,1}\geq \mathcal{H}_{j,2}$, then $\mathbf{r}_{i}^{(1)}=0$, and if $\mathcal{T}_{j,1}< \mathcal{H}_{j,2}$, then $\mathbf{s}_{i}^{(1)}=1$. When using the continuous version, the weights of logical layers are trainable, which means we can choose bounds in an end-to-end way. Since the number of bounds is $2k$ times of features, which could be large, only logical layers with novel logical activation functions are capable of choosing the bounds.
\begin{figure}
    \centering
    \includegraphics[width=0.4\textwidth]{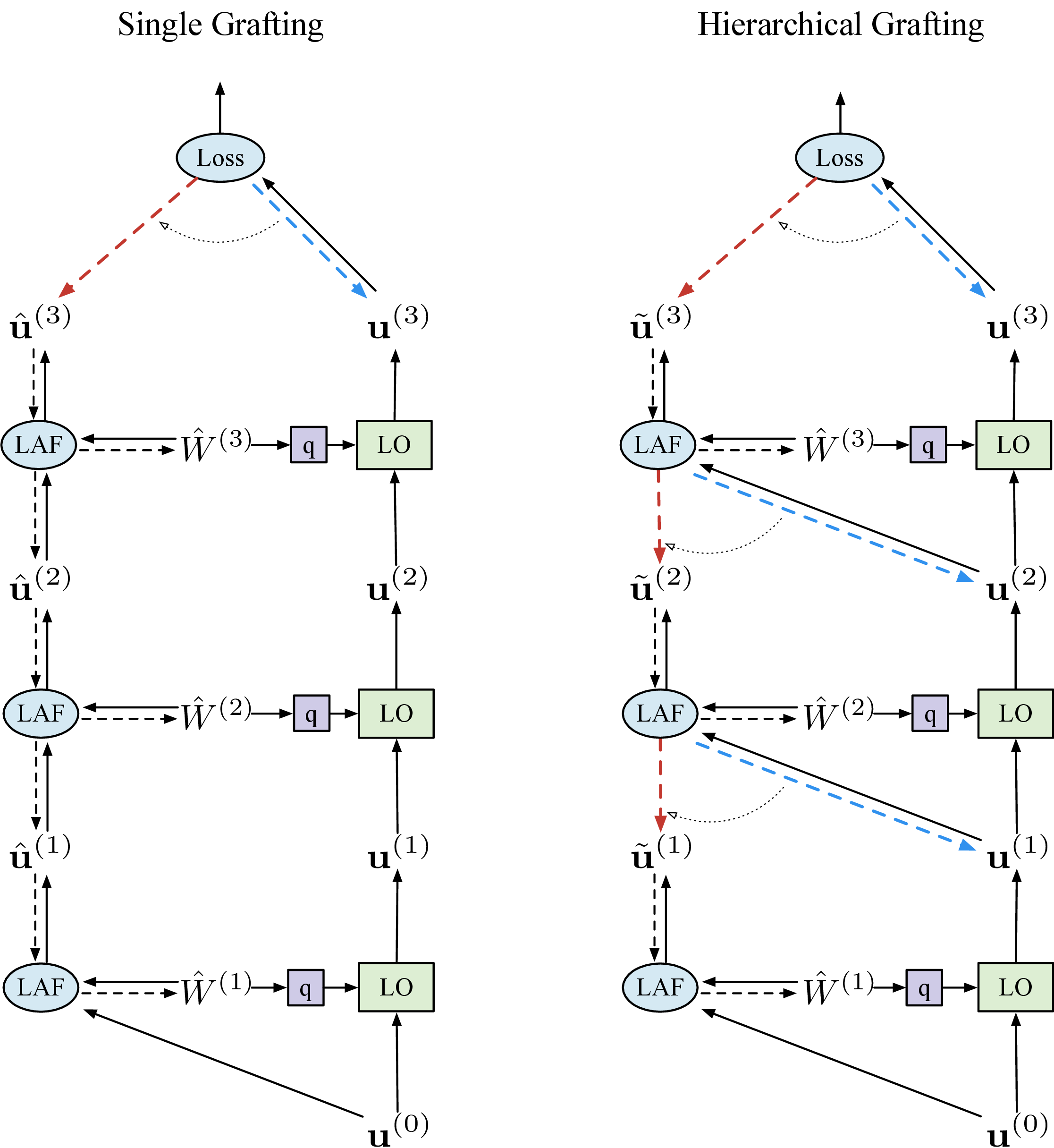}
    \caption{\hll{Simplified computation graphs of Gradient Grafting. The left graph is an example of single grafting, while the right graph is an example of hierarchical grafting for the same layers. Solid arrows with solid lines represent forward pass, while solid arrows with dashed lines represent backpropagation. Each hollow arrow connects a gradient grafting pair. In one pair, the red arrow denotes the grafted gradient, a copy of the gradient represented by the blue arrow. Circles represent differentiable functions, while squares represent non-differentiable functions. After grafting, there exists a backward path from the loss function to all the parameters. We omit the linear layer for better understanding. LAF: Logical Activation Function; LO: Logical Operation; q: quantizer (binarization function).}}
    \label{fig:GradGraftingComparison}
\end{figure}
\subsection{Gradient Grafting}
\label{section:gradient_grafting}
Although RRL can be differentiable with the continuous logical layers, it is challenging to search for a discrete solution in a continuous space \cite{qin2020binary}. One commonly used method to tackle this problem is the Straight-Through Estimator (STE) \cite{courbariaux2016binarized}. The STE method needs gradients at discrete points to update the parameters. However, the gradients of RRL at discrete points have no useful information in most cases (See Appendix \ref{gradients_at_discrete_points}). Therefore STE is not suitable for RRL. Other methods like ProxQuant \cite{bai2018proxquant} and Random Binarization \cite{wang2020transparent} cannot directly optimize for the discrete model and be scalable at the same time.

Inspired by plant grafting, we propose a new training method, called Gradient Grafting, that can effectively train RRL. In stem grafting, one plant is selected for its roots, i.e., rootstock, and the other plant is selected for its stems, i.e., scion. 
\hll{
By stem grafting, we obtain a new plant with the advantages of both two plants. By gradient grafting, we obtain a new backpropagation path that integrates gradient information from both the discrete model and the continuous model. With this backpropagation path, we can directly optimize the loss of the discrete RRL.
}\hll{
\subsubsection{Single Gradient Grafting}
In single Gradient Grafting, the gradient of the loss function w.r.t. the output of the discrete model is the scion, and the gradient of the output of the continuous model w.r.t. the parameters of the continuous model is the rootstock. Specifically, let $\hat{W}^{(l)}$ denote the parameter matrix of the $l$-th continuous layer and $\hat{W}^{(l)}|_{t}$ denote the same matrix at step $t$. Since the $l$-th discrete logical layer shares the same parameter with the $l$-th continuous logical layer, its parameter matrix is $q(\hat{W}^{(l)})$. Here, $q(W)= \1_{W>0.5}$ is the binarization function that binarizes each element of $W$ with 0.5 as the threshold. $\hat{\mathbf{u}}^{(L)}$ and $\mathbf{u}^{(L)}$ are the outputs of the continuous RRL and discrete RRL, respectively. The single Gradient Grafting is formulated by:
\begin{equation}
\frac{\partial \mathcal{L}(\hat{\mathbf{u}}^{(L)})}{\partial \hat{\mathbf{u}}^{(L)}} \leftarrow \frac{\partial \mathcal{L}(\mathbf{u}^{(L)})}{\partial \mathbf{u}^{(L)}},
\end{equation}
where $\leftarrow$ is the grafting operation, and $g_1 \leftarrow g_2$ indicates we can replace the gradient $g_1$ with $g_2$. $\mathcal{L}(\cdot)$ is the loss function.
With single Gradient Grafting, we can update the parameters according to the loss of the discrete RRL as follows:
\begin{equation}
\hat{W}^{(l)}|_{t+1}=\hat{W}^{(l)}|_{t}-\eta \frac{\partial \mathcal{L}(\mathbf{u}^{(L)})}{\partial \mathbf{u}^{(L)}}\cdot \frac{\partial \hat{\mathbf{u}}^{(L)}}{\partial \hat{W}^{(l)}|_{t}},
\end{equation}
where $\eta$ is the learning rate. One simplified computation graph of single Gradient Grafting is shown in the left part of Figure \ref{fig:GradGraftingComparison} for intuitive understanding.
}

\hll{
However, in practice, when the RRL goes deeper (i.e., the number of logical layers increases), the Gradient Grafting is more likely to fail. 
The main reason is that only when the output of the continuous layer is close to the output of its corresponding discrete layer can the gradient information obtained from the continuous layer guide the training. However, during the forward pass, the difference between the outputs of the continuous logical layer and the discrete logical layer is enlarged layer by layer.
Take the left part of Figure \ref{fig:GradGraftingComparison} as an example. The difference between $\hat{\mathbf{u}}^{(1)}$ and $\mathbf{u}^{(1)}$ is caused by the difference between the logical activation function and the logical operation, which is inevitable. This difference further affects the next layer since the input of the next continuous layer is the continuous output $\hat{\mathbf{u}}^{(1)}$ rather than the discrete output $\mathbf{u}^{(1)}$. Therefore, plus the inevitable difference caused by the logical activation function, the difference between $\hat{\mathbf{u}}^{(2)}$ and $\mathbf{u}^{(2)}$ will be further enlarged.
\subsubsection{Hierarchical Gradient Grafting}
\label{sec:hierarchical_gradient_grafting}
One possible solution to avoid the differences caused by multiple layers is changing the forward pass to use the outputs of the discrete layers as the inputs of their next continuous layers. One example is shown in the right part of Figure \ref{fig:GradGraftingComparison}. After changing the forward pass, let $\tilde{\mathbf{u}}^{(l)}$ denote the output of the $l$-th continuous logical layer, then Equation \ref{eq:laf_q} is replaced by:
\begin{equation}
    \tilde{\mathbf{u}}^{(l)}=\textit{LAF}(\mathbf{u}^{(l-1)}, \hat{W}^{(l)})
\end{equation}
}

\hll{
Although the difference caused by the multiple layers is now avoided, with single Gradient Grafting, there is no complete backward path from the loss function to all the parameters since the logical operations are non-differentiable. The example can also be found in the right part of Figure \ref{fig:GradGraftingComparison}. To build a complete backward path to all the parameters, we need to adopt the gradient grafting for each layer, i.e., the hierarchical Gradient Grafting. First, we can observe that the single Gradient Grafting is still applicable, i.e., $\frac{\partial \mathcal{L}(\tilde{\mathbf{u}}^{(L)})}{\partial \tilde{\mathbf{u}}^{(L)}} \leftarrow \frac{\partial \mathcal{L}(\mathbf{u}^{(L)})}{\partial \mathbf{u}^{(L)}}$. Then, for each logical layer, we can graft the gradients as follows:
\begin{equation}
\frac{\partial \tilde{\mathbf{u}}^{(l)}}{\partial \tilde{\mathbf{u}}^{(l-1)}} \leftarrow \frac{\partial \tilde{\mathbf{u}}^{(l)}}{\partial \mathbf{u}^{(l-1)}},\;\;\;\;l\in \{2, 3,\dots, L\}
\end{equation}
After the hierarchical Gradient Grafting, we can obtain a complete backward path from the loss function to all the parameters. For example, we can compute $\frac{\partial \mathcal{L}(\mathbf{u}^{(L)})}{\partial \hat{W}^{(1)}}$ by $\frac{\partial \mathcal{L}(\mathbf{u}^{(L)})}{\partial \mathbf{u}^{(L)}}\cdot \frac{\partial \tilde{\mathbf{u}}^{(L)}}{\partial \mathbf{u}^{(L-1)}}\dots \frac{\partial \tilde{\mathbf{u}}^{(2)}}{\partial \mathbf{u}^{(1)}}\cdot \frac{\partial \tilde{\mathbf{u}}^{(1)}}{\partial \hat{W}^{(1)}}$. The example in the right part of Figure \ref{fig:GradGraftingComparison} can provide more intuition.
}

Gradient Grafting can directly optimize the loss of discrete models and use the gradient information from both continuous and discrete models, which overcomes the problems occurring in RRL training when using other gradient-based discrete model training methods. 
\hll{
Hierarchical Gradient Grafting further enhances its performance on deep RRL.
}
The convergence of Gradient Grafting is verified in the experiments (See Figure \ref{fig:training_loss} and \ref{fig:gradgraft_ablation}).

\subsection{Model Interpretation}
After training with Gradient Grafting, the discrete RRL can be used for testing and interpretation. RRL is easy to interpret, for we can simply consider it as a feature learner and a linear classifier. The binarization layer and logical layers are the feature learner, and they use logical rules to build and describe the new features. The linear classifier, i.e., the linear layer, makes decisions based on the new features. We can first find the important new features by the weights of the linear layer, then understand each new feature by analyzing its corresponding rule. 

One advantage of RRL is that it can be easily adjusted by practitioners to obtain a trade-off between classification accuracy and model complexity. 
Therefore, RRL can satisfy the requirements from different tasks and scenarios. 
There are several ways to limit the model complexity of RRL. First, we can reduce the number of logical layers in RRL, i.e., the depth of RRL, and the number of nodes in each logical layer, i.e., the width of RRL. 
Second, the L1/L2 regularization can be used during training to search for an RRL with shorter rules. The coefficient of the regularization term in the loss function can be considered as a hyperparameter to restrict the model complexity. 
\hll{
Additionally, we find that a trainable temperature of the softmax \cite{hinton2015distilling} in the loss function for classification tasks can also reduce the model complexity, especially when adopted with other regularization \hlr{(see Appendix \ref{appendix:temperature} for the experimental comparisons between trainable temperature and fixed temperature)}.
}
After training, the dead node detection and redundant rule elimination proposed by \cite{wang2020transparent} can also be used for better interpretability.

\begin{table}[!t]
  \caption{Data sets properties.}
  \label{tab:property}
\hll{
  \begin{tabular}{ccccc}
\toprule
Dataset & \#instance & \#class	& \#feature & feature type\\
\midrule	
adni	&	606	&	2	&	49	&	mixed	\\
adult	&	32561	&	2	&	14	&	mixed	\\
bank-marketing	&	45211	&	2	&	16	&	mixed	\\
banknote	&	1372	&	2	&	4	&	continuous	\\
chess	&	28056	&	18	&	6	&	discrete	\\
connect-4	&	67557	&	3	&	42	&	discrete	\\
letRecog	&	20000	&	26	&	16	&	continuous	\\
magic04	&	19020	&	2	&	10	&	continuous	\\
tic-tac-toe	&	958	&	2	&	9	&	discrete	\\
wine	&	178	&	3	&	13	&	continuous	\\
\midrule									
activity	&	10299	&	6	&	561	&	continuous	\\
dota2	&	102944	&	2	&	116	&	discrete	\\
facebook	&	22470	&	4	&	4714	&	discrete	\\
fashion	&	70000	&	10	&	784	&	continuous	\\
      \bottomrule
\end{tabular}
}
\end{table}
\section{Experiments}
In this section, we conduct experiments to evaluate the proposed model and answer the following questions: (\romannumeral1) How is the classification performance of RRL? (\romannumeral2) How is the model complexity of RRL? (\romannumeral3) How is the scalability of the \hll{novel} logical activation functions? (\romannumeral4) How is the convergence of Gradient Grafting compared with other gradient-based discrete model training methods?  
\subsection{Dataset Description and Experimental Settings}
\label{section:dataset_desc_and_exp_settings}
\textbf{Dataset Description.} We took \hll{ten} small and four large public datasets to conduct our experiments, all of which are often used to test classification performance and model interpretability \cite{Dua:2019,xiao2017:online,anguita2013public,rozemberczki2019multiscale,petersen2010alzheimer,wang2022learning}. 
\hll{
Table \ref{tab:property} summarizes the statistics of these 14 data sets.
In Table \ref{tab:property}, the first ten data sets are small data sets, while the last four are large data sets. Discrete or continuous feature type indicates features in that data set are all discrete or all continuous. The mixed feature type indicates the corresponding data set has both discrete and continuous features.
These 14 data sets show the data diversity, ranging from 178 to 102944 instances, from 2 to 26 classes, and from 4 to 4714 original features. See Appendix \ref{appendix:code_and_data} for data sources.
}

\noindent \textbf{Performance Measurement.}
We adopt the F1 score (Macro) as the classification performance metric since some of the data sets are imbalanced, i.e., the numbers of different classes are quite different. We adopt 5-fold cross-validation to evaluate the classification performance more fairly.
The average rank of each model is also adopted for comparisons of classification performance over all the data sets \cite{demvsar2006statistical}.
Considering that reused structures exist in rule-based models, e.g., one branch in Decision Tree can correspond to several rules, we use the total number of edges instead of the total length of all rules as the metric of model complexity for rule-based models. 

\noindent \textbf{Experiment Environment.} We implement our model with PyTorch \cite{paszke2019pytorch}, an open-source machine learning framework. \hll{Experiments are conducted on a Linux server with an Intel Xeon E5 v4 CPU at 2.40GHz and a GeForce RTX 3090 GPU.}

\noindent \textbf{Parameter Settings.} The number of logical layers in RRL ranges from 1 to 4. The number of nodes in logical layers ranges from 32 to 2048, depending on the number of binary features of the data set and the model complexity we need. 
We use the cross-entropy loss during the training. The L2 regularization is adopted to restrict the model complexity, and the coefficient of the regularization term in the loss function is in $\{10^{-2}, 10^{-3},\dots, 10^{-8}, 0\}$.
The numbers of the lower and upper bounds in the binarization layer are both in $\{5, 10, 50\}$. We utilize the Adam \cite{kingma2014adam} method for the training process with a mini-batch size of 32. 
\hll{
The initial learning rate is in $\{5\times 10^{-3},2\times 10^{-3},5\times 10^{-4},2\times 10^{-4},1\times 10^{-4},5\times 10^{-5}\}$.
}On small data sets, RRL is trained for 400 epochs, and we decay the learning rate by a factor of 0.75 every 100 epochs.
On large data sets, RRL is trained for 100 epochs, and we decay the learning rate by a factor of 0.75 every 20 epochs. 
\hll{
For the novel logical activation functions, we adopt $(\alpha, \beta, \gamma)\in \{(0.999, 8, 1), (0.999, 8, 3), (0.9, 3, 3)\}$. 
The initial temperature of the softmax in the loss function is in $\{1,0.1,0.01\}$.
}When parameter tuning is required, 95\% of the training set is used for training and 5\% for validation.

\begin{table*}[!t]
  \caption{5-fold cross validated F1 score (\%) of comparing models on all 14 datasets. $^*$ represents that RRL significantly outperforms all the compared interpretable models (t-test with p $<$ 0.01). The first seven models are interpretable models, while the last \hlr{eight} are complex models.}
  \label{tab:accuracy}
\resizebox{1.\linewidth}{!}{
\hll{
  \begin{tabular}{c|ccccccc|cccccccc}
    \toprule
Dataset	&	\textbf{RRL}	&	C4.5	&	CART	&	SBRL	&	CORESL	&	CRS	&	LR	&	SVM	&	PLNN	&	BNN	&	RF	&	LGBM	&	XGB	&	\hlr{FT}	&	\hlr{SAINT}	\\
\midrule																															
adni	&	84.68	&	79.71	&	81.02	&	80.20	&	81.19	&	83.10	&	83.83	&	83.93	&	84.65	&	83.49	&	83.82	&	\textbf{84.76}	&	84.65	&	82.21	&	84.33	\\
adult	&	80.42	&	77.77	&	77.06	&	79.88	&	70.56	&	\textbf{80.95}	&	78.43	&	79.01	&	79.60	&	77.26	&	79.22	&	80.36	&	80.64	&	79.01	&	79.31	\\
bank-marketing	&	\textbf{77.18}$^*$	&	71.24	&	71.38	&	72.67	&	66.86	&	73.34	&	69.81	&	72.99	&	72.40	&	72.49	&	72.67	&	75.28	&	74.71	&	77.04	&	75.60	\\
banknote	&	\textbf{100.0}$^*$	&	98.45	&	97.85	&	94.44	&	98.49	&	94.93	&	98.82	&	\textbf{100.0}	&	\textbf{100.0}	&	99.64	&	99.40	&	99.48	&	99.55	&	99.93	&	99.04	\\
chess	&	89.73$^*$	&	79.90	&	79.15	&	26.44	&	24.86	&	80.21	&	33.06	&	87.04	&	77.85	&	55.03	&	75.00	&	88.73	&	\textbf{90.04}	&	87.88	&	86.73	\\
connect-4	&	72.01$^*$	&	61.66	&	61.24	&	48.54	&	51.72	&	65.88	&	49.87	&	69.85	&	70.64	&	61.94	&	62.72	&	70.53	&	70.65	&	72.45	&	\textbf{72.85}	\\
letRecog	&	96.14$^*$	&	88.20	&	87.62	&	64.32	&	61.13	&	84.96	&	72.05	&	95.57	&	92.34	&	81.06	&	96.59	&	96.51	&	96.38	&	\textbf{97.17}	&	96.72	\\
magic04	&	86.29$^*$	&	82.44	&	81.20	&	82.52	&	77.37	&	80.87	&	75.72	&	85.35	&	85.50	&	79.50	&	86.48	&	86.67	&	\textbf{86.69}	&	85.95	&	85.46	\\
tic-tac-toe	&	\textbf{100.0}	&	91.70	&	94.21	&	98.39	&	98.92	&	99.77	&	98.12	&	98.07	&	98.26	&	98.92	&	98.37	&	99.89	&	99.89	&	97.84	&	96.95	\\
wine	&	98.37	&	95.48	&	94.39	&	95.84	&	97.43	&	97.78	&	95.16	&	96.05	&	76.07	&	95.77	&	98.31	&	\textbf{98.44}	&	97.78	&	94.63	&	95.50	\\
\midrule																															
activity	&	98.96	&	94.24	&	93.35	&	11.34	&	51.61	&	5.05	&	98.47	&	98.67	&	98.27	&	97.86	&	97.80	&	\textbf{99.41}	&	99.38	&	98.56	&	98.94	\\
dota2	&	\textbf{60.08}$^*$	&	52.08	&	51.91	&	34.83	&	46.21	&	56.31	&	59.34	&	59.25	&	59.46	&	54.76	&	57.39	&	58.81	&	58.53	&	59.70	&	59.58	\\
facebook	&	\textbf{90.11}$^*$	&	80.76	&	81.50	&	31.16	&	34.93	&	11.38	&	88.62	&	87.20	&	89.43	&	85.94	&	87.49	&	85.87	&	88.90	&	86.52	&	88.79	\\
fashion	&	89.64$^*$	&	80.49	&	79.61	&	47.38	&	38.06	&	66.92	&	84.53	&	\textbf{90.09}	&	89.36	&	85.33	&	88.35	&	89.91	&	89.82	&	89.23	&	89.69	\\
\midrule																															
\textbf{AvgRank}	&	\textbf{2.50}	&	11.57	&	12.14	&	11.86	&	12.64	&	9.29	&	10.43	&	6.21	&	6.79	&	9.71	&	7.29	&	3.86	&	3.50	&	\hlr{6.14}	&	\hlr{5.57}	\\
  \bottomrule
\end{tabular}
}
}
\end{table*}
We use validation sets to tune hyperparameters of all the baselines mentioned in Section \ref{section:classification_performance}. We use sklearn to implement Logistic Regression (LR) \cite{kleinbaum2002logistic}, and use the L1 or L2 norm in the penalization. The liblinear is used as the solver. The tolerance for stopping criteria is in \{$10^{-3}$,  $10^{-4}$, $10^{-5}$\}. The inverse of regularization strength is in \{1, 4, 16, 32\}. 
For decision tree, i.e., C4.5 \cite{Quinlan:1993:CPM:583200} or CART \cite{breiman2017classification}, its max depth is in \{None, 5, 10, 20\}. The min number of samples required to split an internal node is in \{2, 8, 16\}, and the min number of samples required to be at a leaf node is in \{1, 8, 16\}. 
For Scalable Bayesian Rule Lists (SBRL) \cite{yang2017scalable}, its $\lambda$ is set to 5 initially, and the min and max rule sizes are set at 1 and 3, respectively. $\eta$ is set to 1, and the numbers of iterations and chains are set to 5000 and 20, respectively. The  minsupport\_pos and minsupport\_neg are set to keep the total number of rules  close to 300.
For Certifiably Optimal Rule Lists (CORELS) \cite{angelino2017learning}, the regularization parameter is in \{0, $10^{-2}$, $10^{-3}$,  $10^{-4}$, $10^{-5}$\}, and the max number of rule lists to search is in \{$10^{4}$, $10^{5}$,  $10^{6}$, $10^{7}$\}. The max cardinality allowed is set to 2 or 3, and the min support rate is in \{0.005, 0.01, 0.025, 0.05\}.
For Concept Rule Sets (CRS), Piecewise Linear Neural Network (PLNN) \cite{chu2018exact}, and \hlr{Binary Neural Network (BNN) \cite{courbariaux2016binarized}}, the candidate sets of learning rate, learning rate decay rate, batch size, model structure (depth and width), and coefficient of the regularization term in the loss function are the same as RRL's. The random binarization rate of CRS is in \{0, 0.7, 0.75, 0.8, 0.85, 0.9, 0.95\}. \hll{We also adopt learning rates in $\{0.02, 0.01, 0.001\}$ for PLNN.
\hlr{For FT~\cite{GorishniyRKB21} and SAINT~\cite{somepalli2021saint}, the embedding size is in $\{8, 16, 32, 64, 128, 256\}$, and the number of stages is in $\{1, 2, 3\}$. We set the number of attention heads in each attention layer to 4. The dropout rate for FT and SAINT is in $\{0.1, 0.2, 0.8\}$.}
For Support Vector Machines (SVM) \cite{scholkopf2001learning}, the linear, RBF and Ploy kernels are used, and we normalize its inputs before training. The tolerance for stopping criteria is in \{$10^{-3}$,  $10^{-4}$, $10^{-5}$\}. The inverse of regularization strength is in $\{2^{-5}, 2^{-4},\dots, 2^{4}, 2^{5}\}$.} The kernel coefficient is set to the reciprocal of the number of features.
For Random Forest \cite{breiman2001random}, its min number of samples required to split an internal node and the min number of samples required to be at a leaf node are the same as the decision tree's. We use LightGBM \cite{ke2017lightgbm} and XGBoost \cite{chen2016xgboost} to implement Gradient Boosted Decision Tree (GBDT). The learning rate of GBDT is in \{0.1, 0.01, 0.001\}, and the max depth of one tree is in \{None, 5, 10, 20\}. 
\hll{
The coefficient of the L2 regularization term is in $\{1.0, 0.1, 0.01, 0\}$.} 
The number of estimators in ensemble models is in \{10, 100, 500\}.
For baselines that cannot directly deal with the continuous value inputs, we use the recursive minimal entropy partitioning algorithm or the KBinsDiscretizer implemented by sklearn to discretize the inputs first.
Grid search is used for the parameter tuning.

\begin{figure*}
    \centering
    \includegraphics[width=0.975\textwidth]{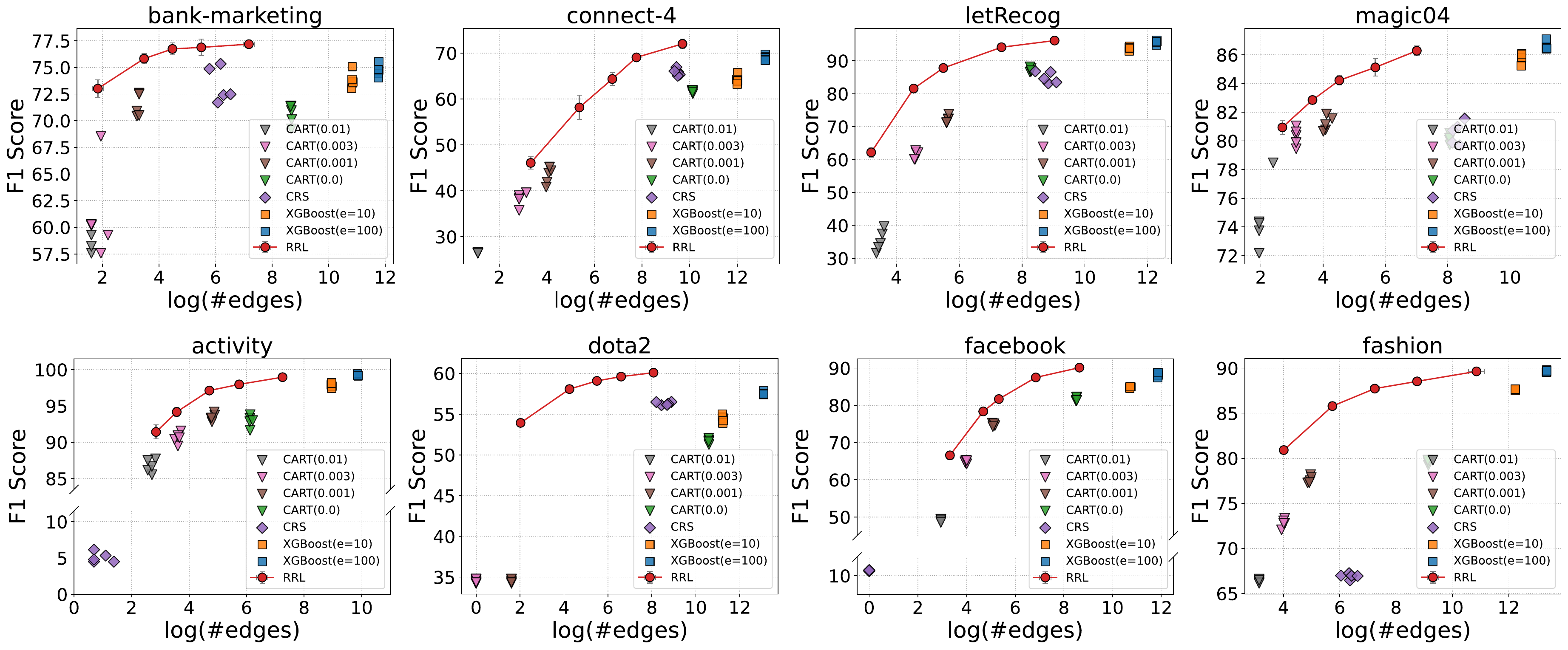}
    \vspace{-4pt}
    \caption{\hll{Scatter plot of F1 score against log(\#edges) for RRL and baselines on eight datasets (see Appendix \ref{appendix:model_complexity} for other datasets). F1 score and log(\#edges) are used to evaluate the classification performance and the model complexity, respectively.}}
    \label{fig:model_complexity}
\end{figure*}
\subsection{Classification Performance}
\label{section:classification_performance}
We compare the classification F1 score (Macro) of RRL with six interpretable models and \hlr{eight} complex models. C4.5 \cite{Quinlan:1993:CPM:583200}, CART \cite{breiman2017classification}, Scalable Bayesian Rule Lists (SBRL) \cite{yang2017scalable}, Certifiably Optimal Rule Lists (CORELS) \cite{angelino2017learning}, and Concept Rule Sets (CRS) \cite{wang2020transparent} are rule-based models. Logistic Regression (LR) \cite{kleinbaum2002logistic} is a linear model. These six models are considered as interpretable models. Piecewise Linear Neural Network (PLNN) \cite{chu2018exact}, Support Vector Machines (SVM) \cite{scholkopf2001learning}, \hlr{Binary Neural Network (BNN) \cite{courbariaux2016binarized}}, Random Forest \cite{breiman2001random}, LightGBM \cite{ke2017lightgbm}, XGBoost \cite{chen2016xgboost}, \hlr{FT~\cite{GorishniyRKB21}, and SAINT~\cite{somepalli2021saint}} are considered as complex models. PLNN is a Multilayer Perceptron (MLP) that adopts piecewise linear activation functions, e.g., ReLU \cite{nair2010rectified}. \hlr{BNN is a special MLP in which each value of the weight and activation tensors is binary. Since we can hardly understand the meaning of its weights and activation values, BNN is not an interpretable model.} RF, LightGBM, and XGBoost are ensemble models. 
\hlr{FT and SAINT are two deep transformer networks recently proposed for tabular data.}

The results are shown in Table \ref{tab:accuracy}, and the first ten data sets are small data sets, while the last four are large data sets. We can observe that RRL performs well on almost all the data sets and gets the best results on 5 data sets. The two-tailed Student’s t-test (p$<$0.01) is used for significance testing, and we can observe that RRL significantly outperforms all the compared interpretable models on 9 out of 14 data sets. The average rank of RRL is also the top among all the models. Only two complex models that use hundreds of estimators, i.e., XGBoost and LightGBM, have comparable results with RRL. 
For Table \ref{tab:accuracy}, the Friedman statistic is \hlr{109.71}, corresponding to a p-value of \hlr{$6.38\times10^{-17}$}. Hence the null hypothesis of no significant differences is rejected. We then calculate pairwise comparisons using Conover post-hoc test, and RRL significantly outperforms all the compared interpretable models with p$<$0.01.
Comparing RRL with LR and other rule-based models, we can see RRL can fit both linear and non-linear data well. CRS performs well on small data sets but fails on large datasets due to the limitation of its logical activation functions and training method. Good results on both small and large data sets verify RRL has good scalability.
Moreover, SBRL and CRS do not perform well on continuous feature data sets like \textit{letRecog} and \textit{magic04} for they need preprocessing to discretize continuous features, which may bring bias to the data sets. On the contrary, RRL overcomes this problem by discretizing features end-to-end.
\hlr{Benefiting from powerful transformers, the complex models FT and SAINT perform well on most datasets. However, due to their complex structure and large number of parameters, FT and SAINT are not suitable for small datasets, e.g., \textit{tic-tac-toe} and \textit{wine}.}
\begin{figure}
    \centering
    \includegraphics[width=0.495\textwidth]{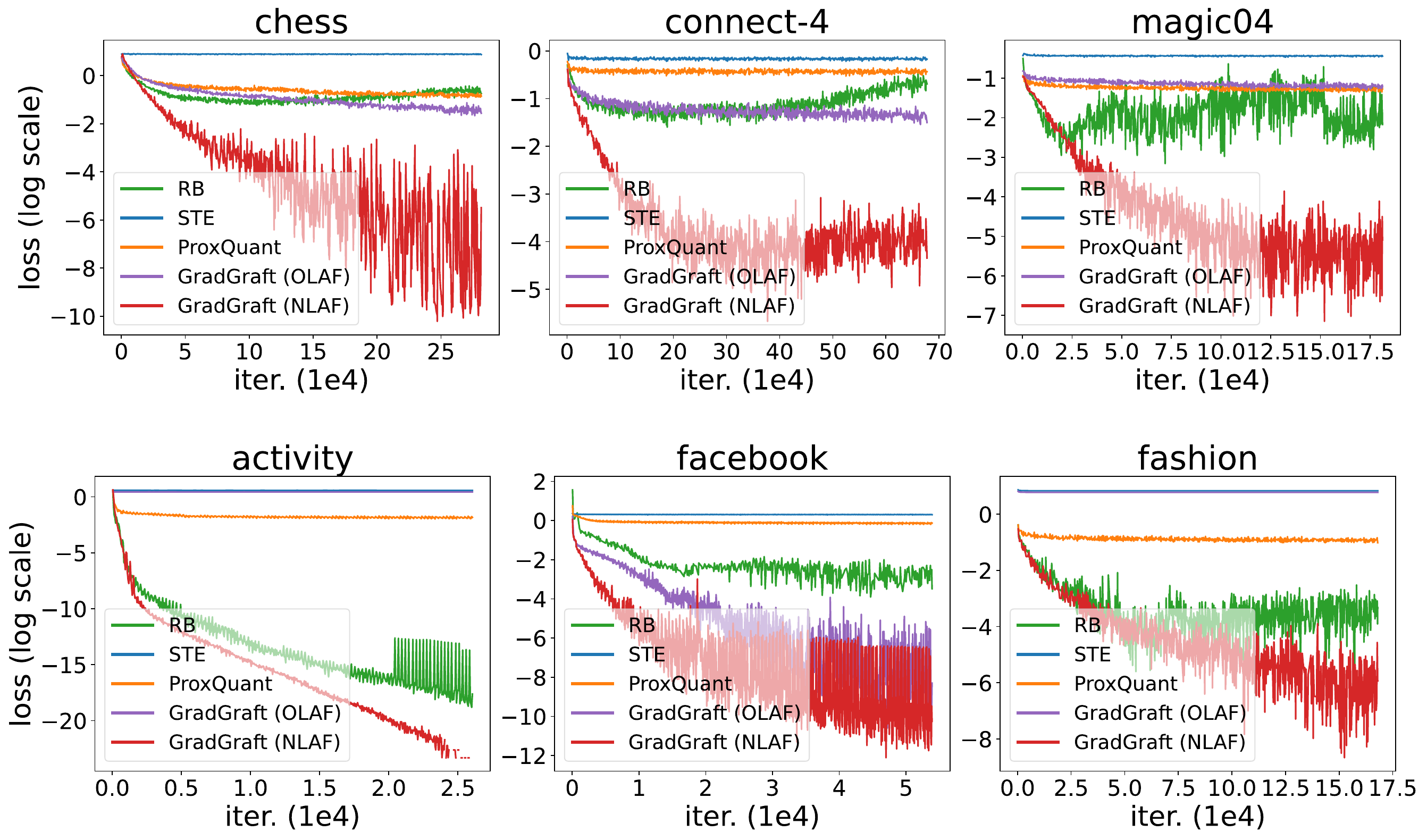}
    \vspace{-15pt}
    \caption{\hll{Training losses of three compared discrete model training methods and Gradient Grafting with original logical activation functions, i.e., GradGraft (OLAF), or novel logical activation functions, i.e., GradGraft (NLAF), on six data sets. The losses are plotted on a log scale for a better viewing experience, and fluctuations at the bottom are actually much smaller than those at the top.}}
    \label{fig:training_loss}
\end{figure}
\subsection{Model Complexity}
Interpretable models seek to keep low model complexity while ensuring high accuracy.  
To show the relationships between accuracy and model complexity of RRL and baselines, we draw scatter plots of F1 score against log(\#edges) for rule-based models in Figure \ref{fig:model_complexity}. (see Appendix \ref{appendix:model_complexity} for other data sets)
The baselines are typical models in different categories of methods with good trade-offs between accuracy and model complexity.
For RRL, the legend markers and error bars indicate means and standard deviations, respectively, of F1 score and log(\#edges) across cross-validation folds.
For baseline models, each point represents an evaluation of one model, on one fold, with one parameter setting.
Therefore, in Figure \ref{fig:model_complexity}, the closer its corresponding points are to the upper left corner, the better one model is.
To obtain RRL with different model complexities, we tune the depth and width of RRL and the coefficient of L2 regularization term. 
The value in CART($\cdot$), e.g., CART(0.03), denotes the complexity parameter used for Minimal Cost-Complexity Pruning \cite{breiman2017classification}, and a higher value corresponds to a simpler tree. We also show results of XGBoost with 10 and 100 estimators.

In Figure \ref{fig:model_complexity}, on both small and large data sets, we can observe that if we connect the results of RRL, it will become a boundary that separating the upper left corner from other models. In other words, if RRL has a close model complexity with one baseline, then the F1 score of RRL will be higher. If the F1 score of RRL is close to one baseline, then the model complexity of RRL will be lower. It indicates that RRL can make better use of rules than rule-based models using heuristic and ensemble methods in most cases.
The results in Figure \ref{fig:model_complexity} also verify that we can adjust the model complexity of RRL by setting the model structure and the coefficient of L2 regularization term. 
In this way, the practitioners are able to select an RRL with suitable classification performance and model complexity for different scenarios, which is crucial for practical applications of interpretable models.

\subsection{Ablation Study}
\noindent \textbf{Novel Logical Activation Functions.} \hll{We compare the training loss of RRL trained by Gradient Grafting with or without novel logical activation functions. The results are shown in Figure \ref{fig:training_loss}, and GradGraft (NLAF) represents RRL using novel logical activation functions, while GradGraft (OLAF) represents RRL using original logical activation functions. We can observe that the original activation functions could converge on small data sets but fail on large data sets, e.g., \textit{activity} and \textit{fashion}, due to the vanishing gradient problem. The novel activation functions overcome this problem and work well on all data sets, which means the novel logical activation functions make RRL more scalable.} It should be noted that GradGraft (OLAF) works well on the large data set \textit{facebook}. The reason is \textit{facebook} is a very sparse data set, and the number of 1 in each binary feature vector is less than 30 (See Appendix \ref{vanishing_gradient_problem} for detailed analyses).

\noindent \textbf{Training Method for Discrete Model.} To show the effectiveness of Gradient Grafting for training RRL, we compare it with other representative gradient-based discrete model training methods, i.e., STE \cite{cour2015bconnect,courbariaux2016binarized}, ProxQuant \cite{bai2018proxquant} and RB \cite{wang2020transparent}, by training RRL (with NLAF) with the same structure. 
\hll{Hyperparameters are set to be the same for each method except the learning rate and exclusive hyperparameters, e.g., random binarization rate for RB, are fine-tuned. Furthermore, if one baseline method cannot converge well with one specific model structure of RRL, we will reduce the depth or the width of RRL to help it converge. The final training losses of the compared discrete model training methods and Gradient Grafting are shown in Figure \ref{fig:training_loss}.
We can see that the convergence of Gradient Grafting, i.e., GradGraft (NLAF), is faster and stabler than other methods on all six data sets.}
As we mentioned in Section \ref{section:gradient_grafting}, RRL has little useful gradient information at discrete points, thus RRL trained by STE cannot converge. Due to the difference between discrete and continuous RRL, RRL trained by ProxQuant and RB cannot converge well as well.

\begin{figure}
    \centering
    \includegraphics[width=0.4\textwidth]{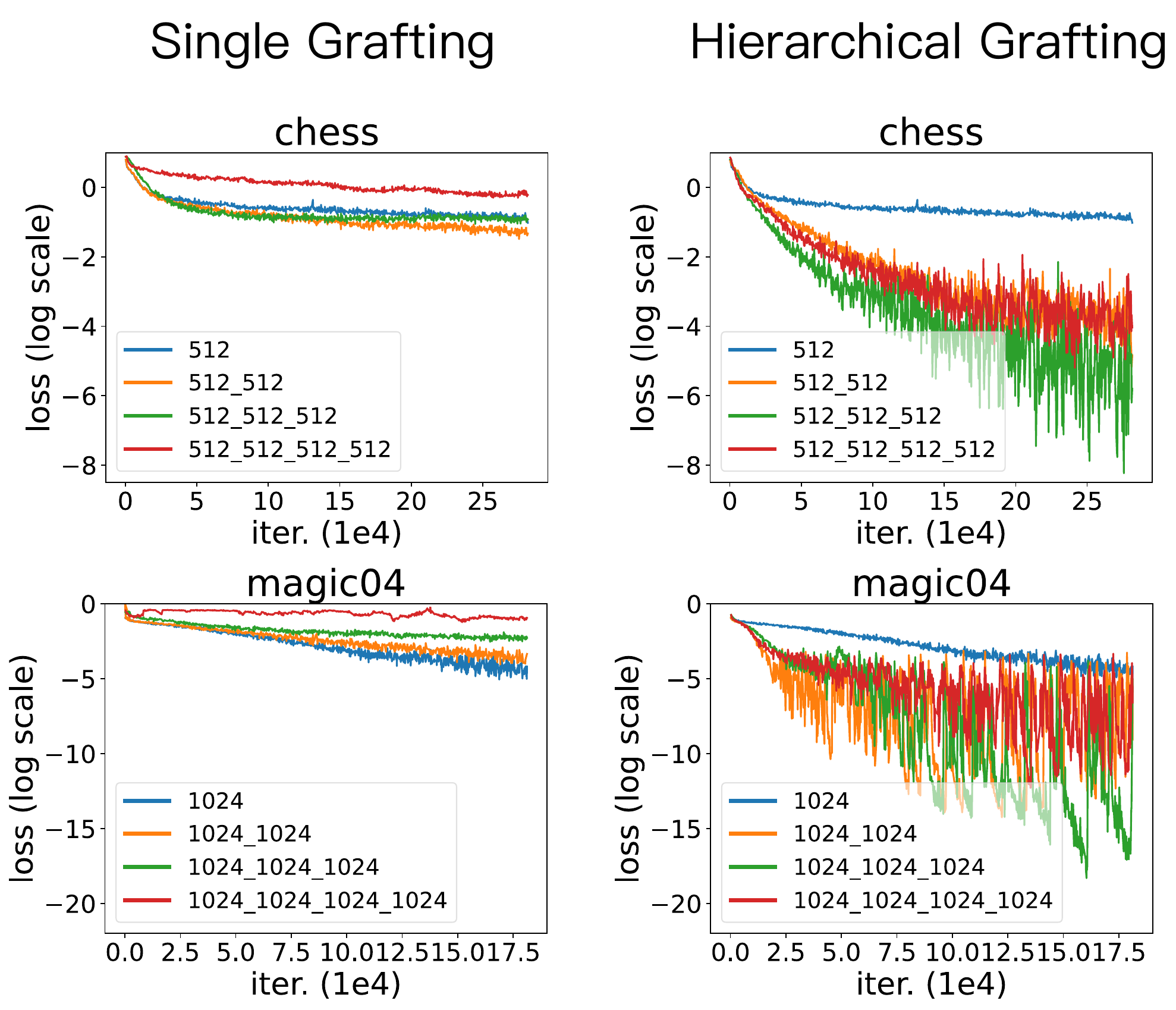}
    \vspace{-10pt}
    \caption{\hll{Training losses of the single and the hierarchical Gradient Grafting on RRL with different depths (i.e., the number of logical layers). The legend labels show the number of nodes in each logical layer. E.g., 512\_512\_512 represents three logical layers, and each layer has 512 nodes. For a better viewing experience, we plot the losses on a log scale. Hence, fluctuations at the bottom are actually much smaller than those at the top.}}
    \label{fig:gradgraft_ablation}
\end{figure}

\noindent \hll{\textbf{Hierarchical Gradient Grafting}. To show how the deep RRL, i.e., RRL with several logical layers, will benefit from the hierarchical Gradient Grafting, we trained RRL with different depths on two data sets using the single and the hierarchical Gradient Grafting, respectively. The results are shown in Figure \ref{fig:gradgraft_ablation}, and the first column shows the single grafting, while the second column shows the hierarchical grafting. 
Comparing these two columns, we can observe that the single Gradient Grafting is more likely to fail when the RRL goes deeper, e.g., RRL with four logical layers cannot converge well using single Gradient Grafting.
On the contrary, hierarchical Gradient Grafting ensures that we can obtain an RRL with better expressivity and capacity by appropriately increasing the number of logical layers. For instance, when using hierarchical Gradient Grafting, RRL with three logical layers converges better on both data sets than RRL with one or two logical layers.
}

\begin{figure}
\centering
  \begin{subfigure}[c]{0.85\linewidth}
    \includegraphics[width=\linewidth]{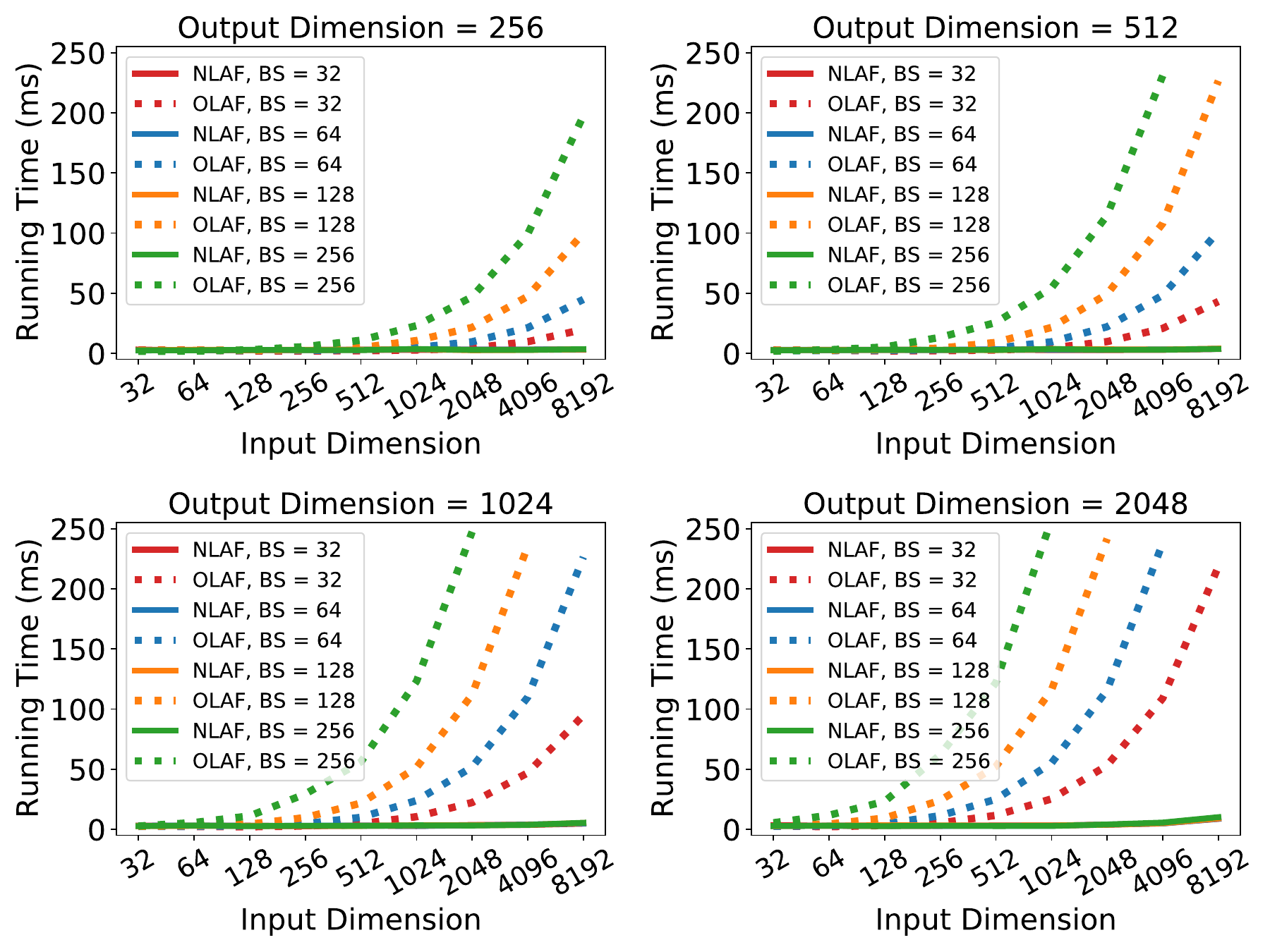}
    \vspace{-15pt}
    \caption{}
    \label{fig:gpu_time}
  \end{subfigure}
  \hfill
  \begin{subfigure}[c]{0.9\linewidth}
    \includegraphics[width=\linewidth]{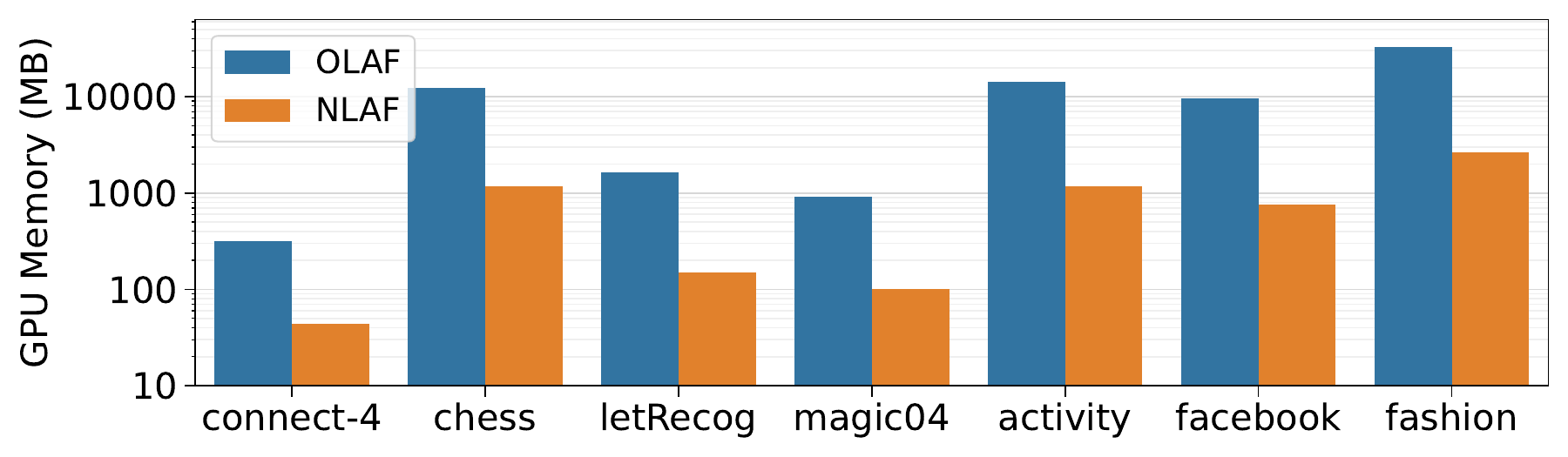}
    \caption{}
    \label{fig:gpu_memory}
  \end{subfigure}
\vspace{-10pt}
\caption{\hll{(a) Time spent on one forward pass and one backpropagation of logical layers with different logical activation functions, input dimensions, output dimensions, and batch sizes (BS). (b) GPU memory usage of RRL (the same model structure as RRL shown in Table \ref{tab:accuracy}) using original or novel logical activation functions.}}
\label{fig:selector}
\end{figure}

\noindent \hll{\textbf{Computation Time and GPU Usage}. 
To show how the novel logical activation functions reduce the computation time and the GPU memory usage compared with the original ones, we first show the time spent on one forward pass and one backward pass of one logical layer in Figure \ref{fig:gpu_time}. We can see that no matter the batch size, input dimension, and output dimension, the novel logical activation functions are faster and more scalable than the original ones.
We then, in Figure \ref{fig:gpu_memory}, show the GPU memory required by RRL trained on different data sets using original or novel logical activation functions (i.e., OLAF or NLAF). The model structures of RRL shown in Figure \ref{fig:gpu_memory} are the same as RRL shown in Table \ref{tab:accuracy}. We can observe that the GPU memory used by NLAF is much less than OLAF due to the elimination of the intermediate result matrix mentioned in Section \ref{sec:NLAF}.}

\begin{figure*}
\centering
  \begin{subfigure}[c]{0.32\linewidth}
    \includegraphics[width=\linewidth]{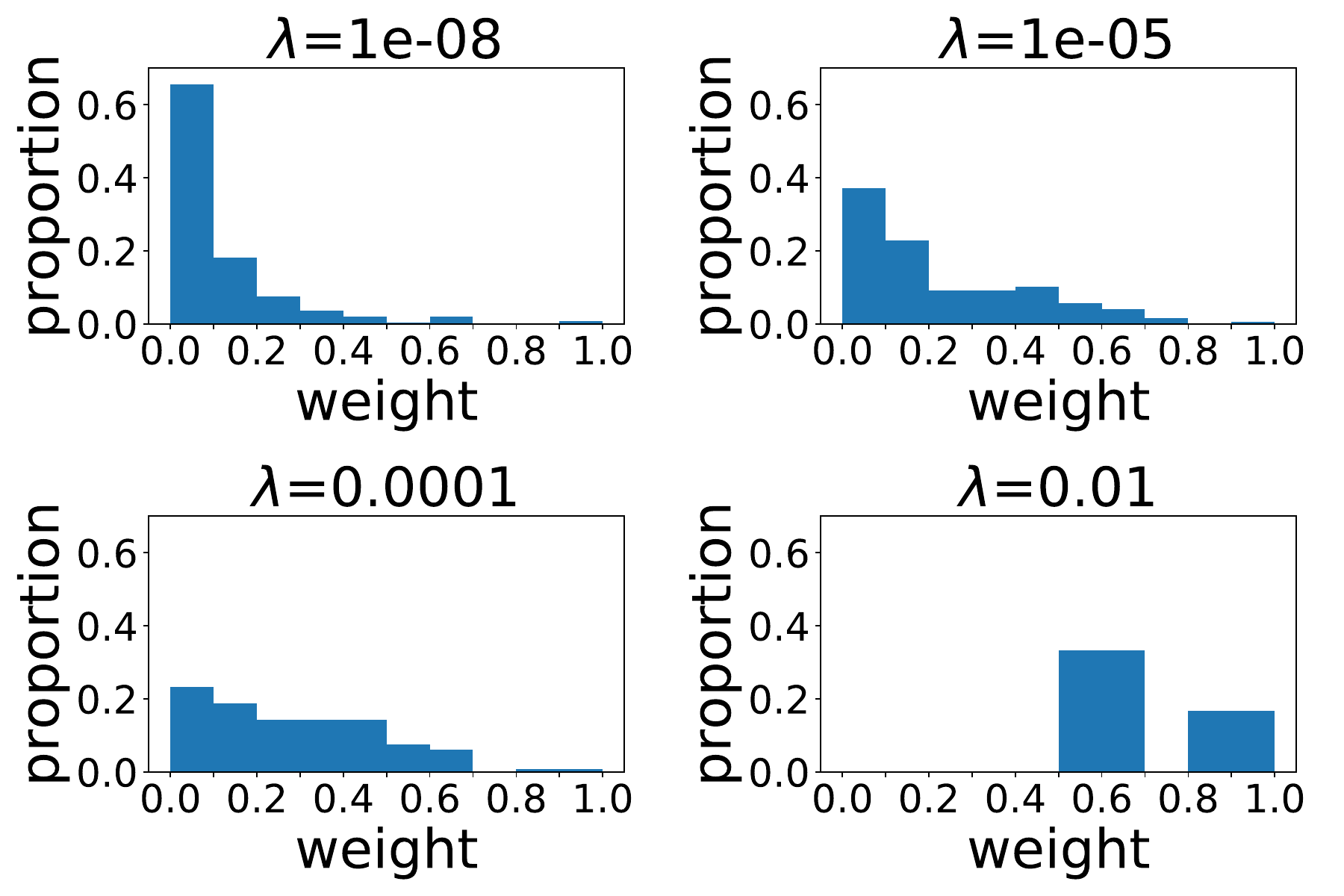}
    \vspace{-15pt}
    \caption{}
    \label{fig:weights_distribution}
  \end{subfigure}
  \hfill
  \begin{subfigure}[c]{0.65\linewidth}
    \includegraphics[width=\linewidth]{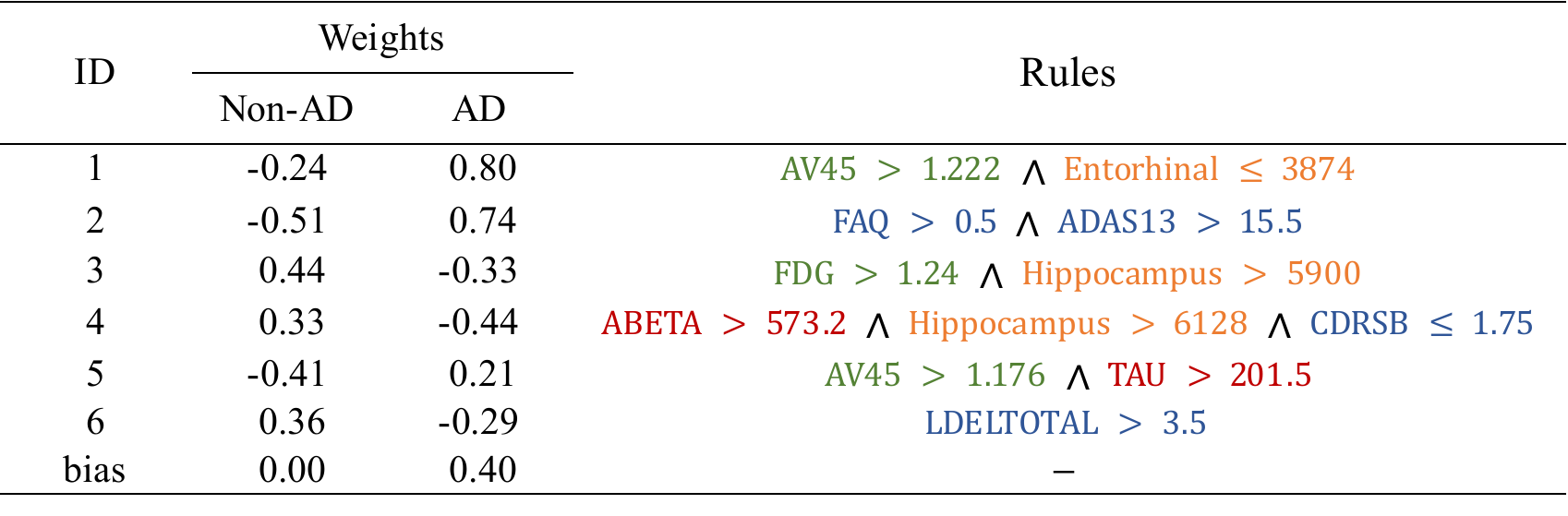}
    \caption{}
    \label{fig:rrl_case_example}
  \end{subfigure}
\vspace{-1pt}
\caption{\hll{(a) The distribution of weights in the linear layer of RRLs trained on the \textit{adni} data set with the same model structure but different $\lambda$, where $\lambda$ is the coefficient of the L2 regularization term. (b) Logical rules obtained from RRL trained on the \textit{adni} data set, \hlr{where Non-AD and AD correspond to different classes}.}}
\label{fig:selector}
\end{figure*}
\subsection{Case Study}
We show what the learned RRL looks like through case studies. Take the \textit{\hll{adni}} data set \cite{petersen2010alzheimer} as an example \hlr{(see Appendix \ref{appendix:case_study_fashion} for case studies on a large data set)}. We first show the distribution of weights in the linear layer of the trained RRLs in Figure \ref{fig:weights_distribution}. 
Each weight in the linear layer corresponds to one rule. 
For a better viewing experience, we show the normalized absolute values of weights.
The model structures of these RRLs are the same, but different coefficients of the L2 regularization term, denoted by $\lambda$, lead to different model complexities.
We can observe that, when $\lambda$ is small, which means RRL is more complex, there are many rules with small weights. These small weight rules are mainly used to slightly adjust the outputs. Hence, they make RRL more accurate but less interpretable. However, in practice, we can ignore these small weight rules and only focus on rules with large weights first. After analyzing rules with large weights and having a better understanding of the learned RRL and the data set, we can then understand those less important rules gradually.
When $\lambda$ is large, the number of rules is small, and we can directly understand the whole RRL rather than understanding RRL step by step.

\hll{
In Figure \ref{fig:rrl_case_example}, we show the whole learned rules of one RRL trained on the \textit{adni} data set (with $\lambda=0.01$). These rules are used to predict if a patient with mild cognitive impairment (MCI) will progress to Alzheimer’s disease (AD) and achieve an F1 score of 0.87. 
\hlr{Each rule has one weight for each class (same for multi-class classification). Those weights tell us how each rule contributes to the final decision on each class.}
Different types of features are marked in different colors, i.e., blue for cognitive test results, orange for Magnetic Resonance Imaging (MRI) results, green for Positron Emission Tomography (PET) results, and red for Cerebrospinal Fluid (CSF) results. 
According to the rules, doctors not only know which features are more important (i.e., automatic feature selection) but also clearly see the cut-off values of these features and know how to combine these features together for diagnosis.
For example, for one patient, if the florbetapir mean of the whole cerebellum (i.e., AV45) is greater than 1.222 and the entorhinal cortex volume (i.e., Entorhinal) is no more than 3874 mm$^3$, then the probability of progressing to AD will be greatly increased according to the weights of the first rule. 
Additionally, these interpretable rules are helpful for doctors to diagnose according to both the model and their domain knowledge.
\hlr{It is worth noting that the reason why RRL can get these neat
and simple rules is that the discrete RRL, used for testing and interpretation, is very sparse, with many weights equal to 0.}
}

\section{Limitations}
\hlrs{Although RRL exhibits good classification performance and model interpretability, it is worth mentioning that RRL still has the following limitations. First, RRL requires feature discretization (binarization) for input continuous features, which is less flexible than complex classification methods like deep learning-based methods.
Second, although the training and inference time of RRL is similar to that of Multilayer Perceptrons (MLP) since their computations are similar, RRL is slightly slower than PLNN and BNN due to the need for more exponential and division operations.
Third, the current design of RRL only considers structured data. When dealing with unstructured data using RRL, it is required to convert them into structured data first. For example, RRL needs to convert a 2D image into a 1D vector, which could lose spatial information.
Lastly, we would like to point out that although RRL achieves much better interpretability over black box models including both ensemble methods and deep learning models, it does not show significant superiority against XGBoost and LightGBM in terms of classification accuracy and training time.
}

\section{Conclusion and Future Work}
\label{section:conclusion}
We propose a new scalable classifier, named Rule-based Representation Learner (RRL), that can automatically learn interpretable rules for data representation and classification. For the particularity of RRL, we propose a new gradient-based discrete model training method, i.e., Gradient Grafting, that directly optimizes the discrete model. We also propose a novel design of logical activation functions to increase the scalability of RRL and make RRL capable of discretizing the continuous features end-to-end. Our experimental results show that RRL enjoys both high classification performance and low model complexity on data sets with different scales. 
\hlr{Additionally, RRL ensures transparency in structure by being a discrete rule-based model. In summary, RRL meets the two main requirements of interpretability, i.e., transparent structure and low complexity.}
\hlrs{In future work, we will explore new forms of rule-based representations to further enhance the flexibility of RRL. In addition, since the current design of RRL is limited to structured data, we will also extend RRL to suit more unstructured data in future work.}

\ifCLASSOPTIONcompsoc
  \section*{Acknowledgments}
\else
  \section*{Acknowledgment}
\fi

This work was supported in part by National Key Research and Development Program of China under Grant No. 2020YFA0804503, National Natural Science Foundation of China under Grant No. 92270119, 62272264, and 62072182, Beijing Academy of Artificial Intelligence (BAAI), and Shandong Provincial Natural Science Foundation(ZR2022QF114). A preliminary conference version of this paper appeared in NeurIPS 2021.

\ifCLASSOPTIONcaptionsoff
  \newpage
\fi



\bibliographystyle{IEEEtran}
\bibliography{IEEEabrv, IEEEexample}
%



%

\begin{IEEEbiography}[{\includegraphics[width=1.1in,height=1.35in,clip,keepaspectratio]{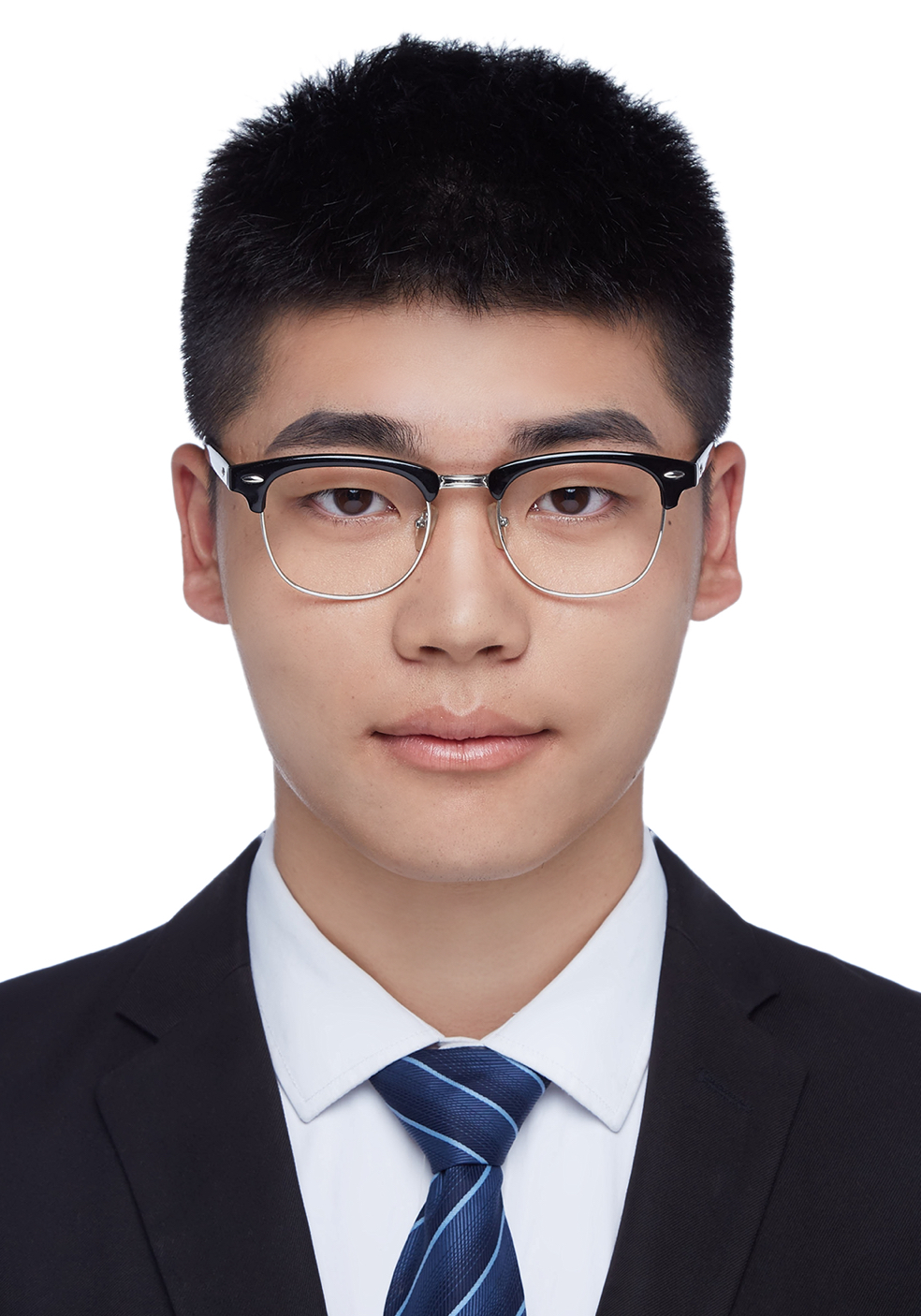}}]{Zhuo Wang}
received his Ph.D. degree in computer science and technology from Tsinghua University, Beijing, China, in 2023. He is currently an algorithm engineer with Ant Group. His research interests mainly include interpretable machine learning models and their financial and medical applications.
\end{IEEEbiography}

\begin{IEEEbiography}[{\includegraphics[width=1.1in,height=1.35in,clip,keepaspectratio]{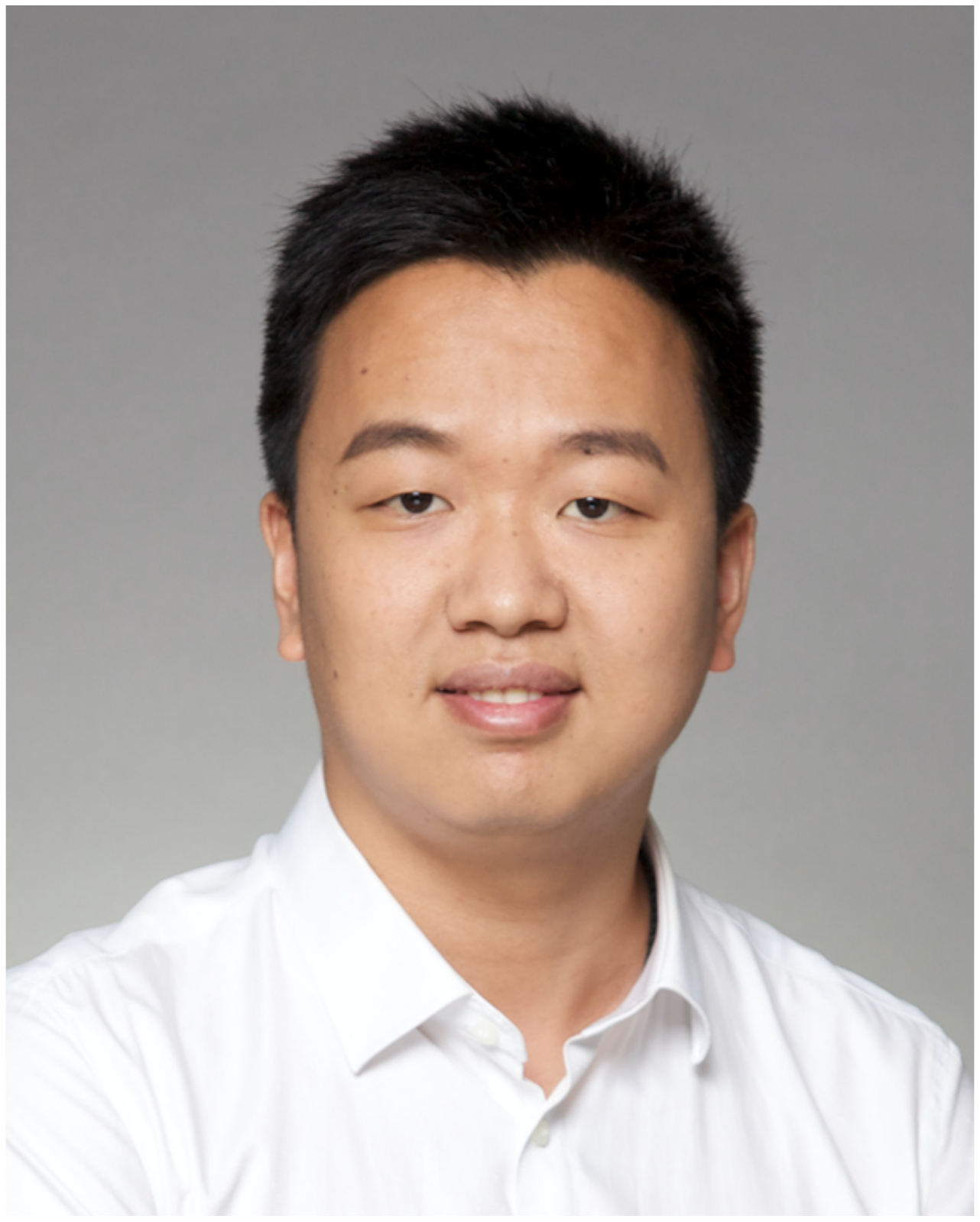}}]{Wei Zhang} received his Ph.D. degree in computer science and technology from Tsinghua university, Beijing, China, in 2016. He is currently a professor in the School of Computer Science and Technology, East China Normal University, Shanghai, China.
His research interests mainly include user data mining and machine learning applications.
He is a senior member of China Computer Federation.
\end{IEEEbiography}


\begin{IEEEbiography}[{\includegraphics[width=1.1in,height=1.35in,clip,keepaspectratio]{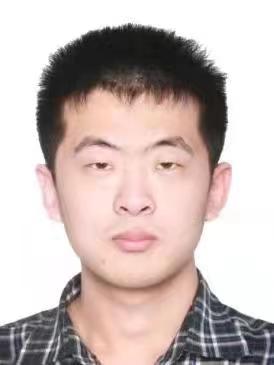}}]{Ning Liu}
received his Ph.D. degree in computer science and technology from Tsinghua university, Beijing, China, in 2021. He is currently an assistant professor in the School of Software, Shandong University, Jinan, China. His research interests mainly include data mining and knowledge-driven applications, especially textural data and sequential data mining. He now is a member of China Computer Federation.
\end{IEEEbiography}

\begin{IEEEbiography}[{\includegraphics[width=1.1in,height=1.35in,clip,keepaspectratio]{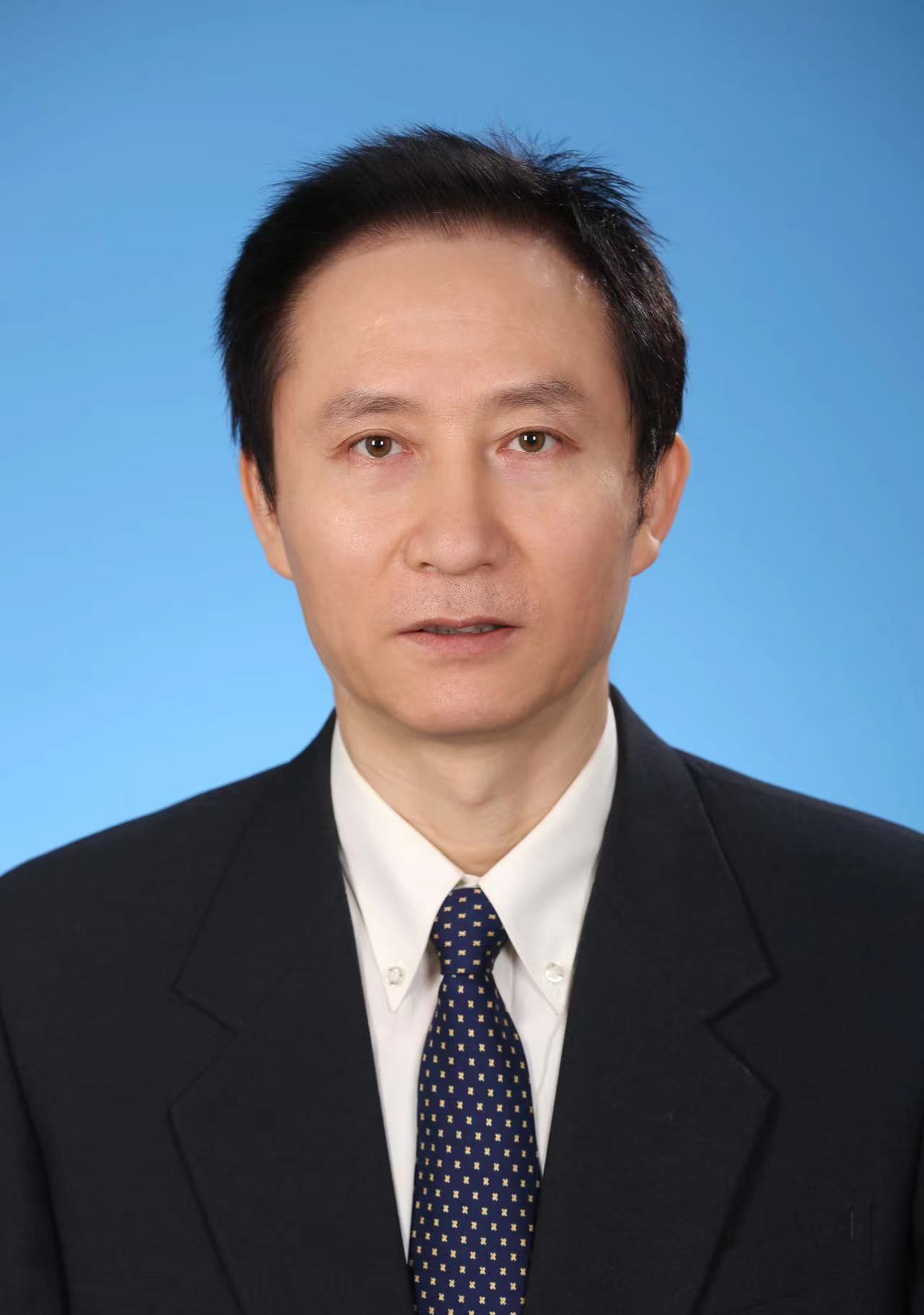}}]{Jianyong Wang} received the PhD degree in computer science from the Institute of Computing Technology, Chinese Academy of Sciences, in 1999. He is currently a tenured professor in the Department of Computer Science and Technology, Tsinghua University, Beijing, China, and with the Jiangsu Collaborative Innovation Center for Language Ability, Jiangsu Normal University, Xuzhou, China. His research interests mainly include data mining and knowledge discovery. He has served as a principal investigator at Beijing Academy of Artificial Intelligence (BAAI), the deputy director of academic committee of Jiangsu Key Laboratory of Big Data Security and Intelligent Processing, conference program co-chair of IEEE ICDM’19, WISE’15, BioMedCom'14, WAIM’13, ADMA’11, and NDBC’10, and an associate editor of the IEEE Transactions on Knowledge and Data Engineering, the ACM Transactions on Knowledge Discovery from Data, Journal of Alzheimer's Disease, and Journal of Personalized Medicine. He has received the best posters award at the WWW 2008 conference, the 2009 and 2010 annual HP Labs Innovation Research awards. Dr. Wang is a recipient of the 2007 annual Program for New Century Excellent Talents in University of Ministry of Education of China, the 2009 annual Okawa Foundation Research Grant, 2017 annual IEEE Fellow, and 2017 annual Fellow of Chinese Association for Artificial Intelligence (CAAI).
\end{IEEEbiography}



\clearpage
\appendices
\section{Data Source}
\label{appendix:code_and_data}

The datasets used in this paper come from the UCI machine learning repository, GitHub, and ADNI. The links to the datasets are: 
\textit{adni}\footnote{\url{https://adni.loni.usc.edu/}}, \textit{adult}\footnote{\url{https://archive.ics.uci.edu/ml/datasets/Adult}}, \textit{bank-marketing}\footnote{\url{http://archive.ics.uci.edu/ml/datasets/Bank+Marketing}}, \textit{banknote}\footnote{\url{https://archive.ics.uci.edu/ml/datasets/banknote+authentication}}, \textit{chess}\footnote{\url{https://archive.ics.uci.edu/ml/datasets/Chess+\%28King-Rook+vs.+King\%29}}, \textit{connect-4}\footnote{\url{http://archive.ics.uci.edu/ml/datasets/connect-4}}, \textit{letRecog}\footnote{\url{https://archive.ics.uci.edu/ml/datasets/Letter+Recognition}}, \textit{magic04}\footnote{\url{https://archive.ics.uci.edu/ml/datasets/magic+gamma+telescope}}, \textit{tic-tac-toe}\footnote{\url{https://archive.ics.uci.edu/ml/datasets/Tic-Tac-Toe+Endgame}}, \textit{wine}\footnote{\url{https://archive.ics.uci.edu/ml/datasets/wine}}, \textit{activity}\footnote{\url{https://archive.ics.uci.edu/ml/datasets/human+activity+recognition+using+smartphones}}, \textit{dota2}\footnote{\url{https://archive.ics.uci.edu/ml/datasets/Dota2+Games+Results}}, \textit{facebook}\footnote{\url{https://archive.ics.uci.edu/ml/datasets/Facebook+Large+Page-Page+Network}}, \textit{fashion}\footnote{\url{https://github.com/zalandoresearch/fashion-mnist}}. 

\section{Vanishing Gradient Problem}
\label{vanishing_gradient_problem}
The partial derivative of each node in $\hat{\mathcal{S}}^{(l)}$ w.r.t. its directly connected weights and w.r.t. its directly connected nodes are given by:
\begin{equation}
\label{eq:derivative_sw}
\frac{\partial \hat{\mathbf{s}}_{i}^{(l)}}{\partial \hat{W}_{i,j}^{(l,1)}}
=\hat{\mathbf{u}}_{j}^{(l-1)} \cdot \prod_{k \neq j}(1-F_{d}(\hat{\mathbf{u}}_{k}^{(l-1)}, \hat{W}_{i,k}^{(l,1)}))
\end{equation}
\begin{equation}
\label{eq:derivative_su}
\frac{\partial \hat{\mathbf{s}}_{i}^{(l)}}{\partial \hat{\mathbf{u}}_{j}^{(l-1)}}
=\hat{W}_{i,j}^{(l,1)} \cdot \prod_{k \neq j}(1-F_{d}(\hat{\mathbf{u}}_{k}^{(l-1)}, \hat{W}_{i,k}^{(l,1)}))
\end{equation}
Similar to the analysis of Equation \ref{eq:derivative_rw_ru}, due to $\hat{\mathbf{u}}_{k}^{(l-1)}$ and $\hat{W}_{i,k}^{(l,1)}$ are in the range $[0, 1]$, the values of $(1-F_d(\cdot))$ in Equation \ref{eq:derivative_sw} and \ref{eq:derivative_su} are in the range $[0, 1]$ as well. If $\mathbf{n}_{l-1}$ is large and most of the values of $(1-F_d(\cdot))$ are not 1, then the derivative is close to 0 because of the multiplications. Therefore, both the conjunction function and the disjunction function suffer from the vanishing gradient problem.

If the large data set is very sparse and the number of 1 in each binary feature vector (for RRL the binary feature vector is $\mathbf{u}^{(0)}$) is less than about one hundred, there will be no vanishing gradient problem for nodes in $\hat{\mathcal{S}}^{(1)}$. The reason is when the number of 1 in each feature vector is less than about one hundred, in Equation \ref{eq:derivative_sw} and \ref{eq:derivative_su}, most of the values of $(1-F_d(\cdot))$ are 1, and only less than one hundred values of $(1-F_d(\cdot))$ are not 1, then the result of the multiplication is not very close to 0. The \textit{facebook} data set is an example of this case. However, if the number of 1 in each binary feature vector is more than about one hundred, the vanishing gradient problem comes again.
\section{Gradients at Discrete Points}
\label{gradients_at_discrete_points}
The gradients of RRL with original logical activation functions at discrete points can be obtained by Equation \ref{eq:derivative_rw_ru}, \ref{eq:derivative_sw} and \ref{eq:derivative_su}. Take Equation \ref{eq:derivative_sw} as an example, discrete points mean all the weights of logical layers are 0 or 1, which also means the values of all the nodes in $\hat{\mathcal{U}}^{(l)}$ are 0 or 1, $l\in\{0,1,\dots,L-1\}$. Hence, in Equation \ref{eq:derivative_sw}, $\hat{\mathbf{u}}_{j}^{(l-1)}, (1-F_d(\cdot)) \in \{0,1\}$, and the whole equation is actually multiplications of several 0 and several 1. Only when $\hat{\mathbf{u}}_{j}^{(l-1)}$ and $(1-F_d(\cdot))$ are all 1, the derivative in Equation \ref{eq:derivative_sw} is 1, otherwise, the derivative is 0. Therefore, the gradients at discrete points have no useful information in most cases. The analyses of Equation \ref{eq:derivative_rw_ru} and \ref{eq:derivative_su} are similar.

\hll{
To analyze the gradients of RRL with novel logical activation functions at discrete points, we first calculate the partial derivative of Equation \ref{eq:basic_laf_mm} w.r.t. one of its inputs: 
\begin{equation}
\label{eq:derivative_nlaf_hj}
\frac{\partial \mathcal{Q}(\mathbf{h}, W_{i})}{\partial \mathbf{h}_{j}}=\gamma \cdot \mathcal{Q}(\mathbf{h}, W_{i})^{\frac{\gamma+1}{\gamma}} \cdot \mathcal{G}(W_{i,j}) \cdot \frac{\partial \mathcal{G}(\mathbf{h}_{j})}{\partial \mathbf{h}_{j}}
\end{equation}
At discrete points, $W_{i,j}, \mathbf{h}_{j}\in\{0, 1\}$. When $W_{i,j}=0$, $\mathcal{G}(W_{i,j})=0$. When $\mathbf{h}_{j}=0$, $\frac{\partial \mathcal{G}(\mathbf{h}_{j})}{\partial \mathbf{h}_{j}}=\frac{-\alpha^{\beta}\beta\mathbf{h}_{j}^{\beta-1}}{(1-(\alpha\mathbf{h}_{j})^{\beta})^{2}}=0$. In other words, if $W_{i,j}$ or $\mathbf{h}_{j}$ equals 0, then $\frac{\partial \mathcal{Q}(\mathbf{h}, W_{i})}{\partial \mathbf{h}_{j}}=0$. When $W_{i,j}=\mathbf{h}_{j}=1$, according to Equation \ref{eq:basic_laf_mm} and Table \ref{tab:truth_table}, $\mathcal{Q}(\mathbf{h}, W_{i})$ is close to 0, which means $\frac{\partial \mathcal{Q}(\mathbf{h}, W_{i})}{\partial \mathbf{h}_{j}}$ is also close to 0.
Hence, at discrete points, $\frac{\partial \mathcal{Q}(\mathbf{h}, W_{i})}{\partial \mathbf{h}_{j}}$ is close to 0 in most cases, and the analyses of $\frac{\partial \mathcal{Q}(\mathbf{h}, W_{i})}{\partial W_{i,j}}$ are similar. Therefore, the gradients at discrete points have little useful information. 
}

\begin{figure}[t!]
    \centering
    \includegraphics[width=0.47\textwidth]{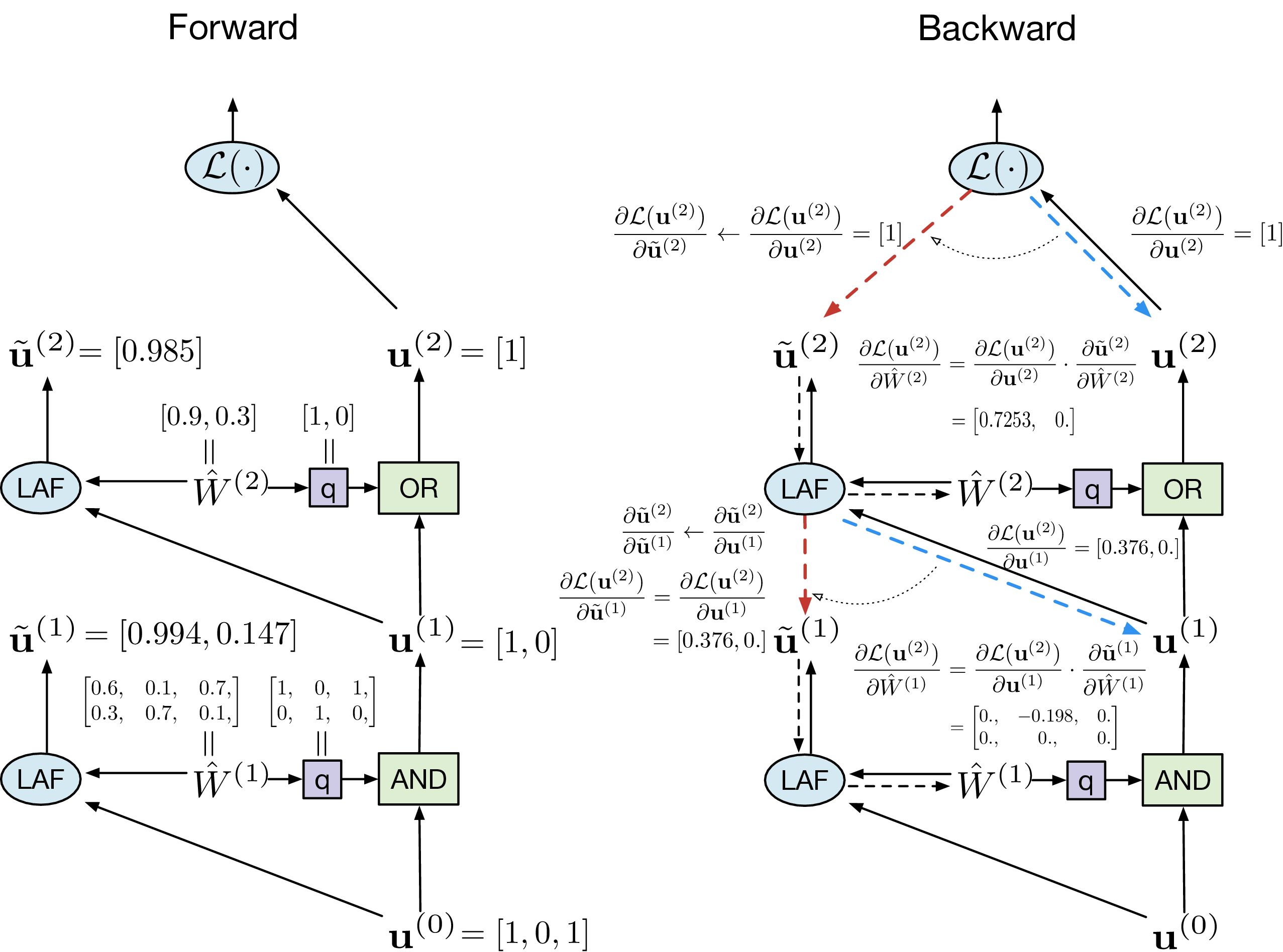}
    \caption{\hlr{One example of the Hierarchical Gradient Grafting. The left part shows the forward pass while the right part shows the backward pass.}}
    \label{fig:gradient_grafting_example}
\end{figure}

\hlr{
\section{Gradient Grafting Example}
\label{appendix:gradient_grafting_example}
}
\hlr{
To get more intuition about the Hierarchical Gradient Grafting, we provide an example in Figure \ref{fig:gradient_grafting_example}. To simplify the whole process, we only use two logical layers, and the first logical layer only has a conjunction layer while the second logical layer only has a disjunction layer. In Figure \ref{fig:gradient_grafting_example}, the left part shows the forward pass while the right part shows the backward pass. Let us begin with the forward pass. The input is $\mathbf{u}^{(0)}=[1,0,1]$. For the first logical layer, its weights are $\hat{W}^{(1)}=\begin{bmatrix} 0.6, & 0.1, & 0.7,\\ 0.3, & 0.7, & 0.1,\end{bmatrix}$, and $q(\hat{W}^{(1)})=\begin{bmatrix}1, & 0, & 1,\\ 0, & 1, & 0,\end{bmatrix}$.
}
\hlr{
Since the first logical layer is a conjunction layer, $\mathbf{u}^{(1)}_{1} = \mathbf{u}^{(0)}_{1}\wedge\mathbf{u}^{(0)}_{3}=1$ and $\mathbf{u}^{(1)}_{2} = \mathbf{u}^{(0)}_{2}=0$. According to the novel LAF (we set all $(\alpha, \beta, \gamma)=(0.9,3,3)$), we can get $\tilde{\mathbf{u}}^{(1)}=[0.994, 0.147]$. As we mentioned in Section \ref{sec:hierarchical_gradient_grafting}, to avoid the difference between the outputs of the continuous logical layer and the discrete logical layer being enlarged layer by layer, we only use $\mathbf{u}^{(1)}$ as the input of the second layer. Similar to the first logical layer, we can get $\mathbf{u}^{(2)}=[1]$ and $\tilde{\mathbf{u}}^{(2)}=[0.985]$. Since we want to optimize the discrete RRL, we input $\mathbf{u}^{(2)}$ into the loss function $\mathcal{L}(\cdot)$. During the forward pass, $\tilde{\mathbf{u}}^{(1)}$ and $\tilde{\mathbf{u}}^{(2)}$ seem useless. In fact, they could be very useful during the backward pass.
}

\hlr{
After the forward pass, we can now do the backward pass. We take $\frac{\partial \mathcal{L}(\mathbf{u}^{(2)})}{\partial \mathbf{u}^{(2)}}=[1]$ as an example. Since the logical operations are non-differentiable, we cannot find a complete backward path from $\mathcal{L}(\cdot)$ to $\hat{W}^{(2)}$. Therefore, we need Gradient Grafting to build a complete backward path. 
According to the gradient grafting $\frac{\partial \mathcal{L}(\mathbf{u}^{(2)})}{\partial \tilde{\mathbf{u}}^{(2)}}\leftarrow\frac{\partial \mathcal{L}(\mathbf{u}^{(2)})}{\partial \mathbf{u}^{(2)}}$, we can get $\frac{\partial \mathcal{L}(\mathbf{u}^{(2)})}{\partial \tilde{\mathbf{u}}^{(2)}}=\frac{\partial \mathcal{L}(\mathbf{u}^{(2)})}{\partial \mathbf{u}^{(2)}}=[1]$. Then, we can calculate $\frac{\partial \mathcal{L}(\mathbf{u}^{(2)})}{\partial \hat{W}^{(2)}}$ by the following equation:
\begin{equation}
    \frac{\partial \mathcal{L}(\mathbf{u}^{(2)})}{\partial \hat{W}^{(2)}} = \frac{\partial \mathcal{L}(\mathbf{u}^{(2)})}{\partial \tilde{\mathbf{u}}^{(2)}}\cdot \frac{\partial \tilde{\mathbf{u}}^{(2)}}{\partial \hat{W}^{(2)}} = \frac{\partial \mathcal{L}(\mathbf{u}^{(2)})}{\partial \mathbf{u}^{(2)}}\cdot \frac{\partial \tilde{\mathbf{u}}^{(2)}}{\partial \hat{W}^{(2)}}.
\end{equation}
Now, we can update $\hat{W}^{(2)}$ by $\frac{\partial \mathcal{L}(\mathbf{u}^{(2)})}{\partial \hat{W}^{(2)}}$.
Similarly, for the first layer, we can get $\frac{\partial \mathcal{L}(\mathbf{u}^{(2)})}{\partial \mathbf{u}^{(1)}}$ since the LAF is differentiable. However, we need $\frac{\partial \mathcal{L}(\mathbf{u}^{(2)})}{\partial \tilde{\mathbf{u}}^{(1)}}$ to get $\frac{\partial \mathcal{L}(\mathbf{u}^{(2)})}{\partial \hat{W}^{(1)}}$. Therefore, we do the gradient grafting $\frac{\partial \tilde{\mathbf{u}}^{(2)}}{\partial \tilde{\mathbf{u}}^{(1)}} \leftarrow \frac{\partial \tilde{\mathbf{u}}^{(2)}}{\partial \mathbf{u}^{(1)}}$. After grafting, we can get $\frac{\partial \mathcal{L}(\mathbf{u}^{(2)})}{\partial \hat{W}^{(1)}}$ by:
\begin{equation}
    \frac{\partial \mathcal{L}(\mathbf{u}^{(2)})}{\partial \hat{W}^{(1)}} = \frac{\partial \mathcal{L}(\mathbf{u}^{(2)})}{\partial \tilde{\mathbf{u}}^{(1)}}\cdot \frac{\partial \tilde{\mathbf{u}}^{(1)}}{\partial \hat{W}^{(1)}} = \frac{\partial \mathcal{L}(\mathbf{u}^{(2)})}{\partial \mathbf{u}^{(1)}}\cdot \frac{\partial \tilde{\mathbf{u}}^{(1)}}{\partial \hat{W}^{(1)}}
\end{equation}
Now, we can also update $\hat{W}^{(1)}$, and all the parameters are updated.
}

\hlr{
\section{Training and Inference Time}
\label{appendix:training_inference_time}
To show how fast RRL is compared with other representative baselines, we show their training time and inference time in Table \ref{tab:training_time} and Table \ref{tab:inference_time}, respectively. In the tables, the first five models are Neural Network-based (NN-based) models, while the last six models are non-NN-based models. 
To compare NN-based models more fairly, we show the time required to train for 100 epochs in Table \ref{tab:training_time}. 
}

\hlr{
We can see that the training and inference time of RRL is similar to neural networks like Multilayer Perceptrons (MLP), e.g., PLNN and BNN, since their computations are quite similar. 
However, due to the need for more exponential and division operations in RRL, RRL is slightly slower than PLNN and BNN, but this is still acceptable. Compared with RRL, PLNN, and BNN, the two deep transformer networks (i.e., FT and SAINT) are very slow for having a very complex model structure and too many parameters. 
}

\hlr{
We can also see that, compared with NN-based models, the non-NN-based models (especially the tree-based models) need less training time, since they do not need the time-consuming gradient descent method to train themselves. To obtain better performance, ensemble models (e.g., XGBoost and LightGBM) use hundreds of estimators, so they need more training and inference time than a single decision tree.
If we compare RRL with XGBoost, we can see that benefiting from parallel tree structure boosting and efficient cacheable block structure, the latest official version of XGBoost is faster than RRL on most datasets. However, we can also observe that, for datasets with many features, e.g., facebook (\#feature=4714), RRL is faster than XGBoost on both training and inference.
}

\begin{table*}[!t]
\centering
  \caption{\hlr{Training time.}}
  \label{tab:training_time}
\resizebox{1.\linewidth}{!}{
\hlr{
  \begin{tabular}{c|ccccc|cccccc}
    \toprule
Dataset & RRL & PLNN  & BNN & SAINT & FT  & CART  & LR  & SVM & RF  & XGBoost & LightGBM\\
\midrule
adni  & 11s & 7s  & 6s  & 3m 12s  & 2m 32s  & 15ms  & 434ms & 102ms & 1s  & 2s  & 1s\\
adult & 7m 35s  & 4m 20s  & 4m 37s  & 14m 36s & 10m 2s  & 346ms & 128ms & 10m 23s & 14s & 20s & 9s\\
bank-marketing  & 10m 34s & 6m 1s & 5m 56s  & 31m 44s & 13m 32s & 354ms & 788ms & 31s & 19s & 18s & 10s\\
banknote  & 24s & 14s & 12s & 1m 54s  & 2m 34s  & 2.9ms & 7.9ms & 28ms  & 972ms & 2s  & 1s\\
\midrule
activity  & 4m 22s  & 1m 37s  & 1m 38s  & 1h 26m 52s  & 27m 32s & 5s  & 7s  & 36s & 59s & 26s & 18s\\
dota2 & 27m 10s & 14m 30s & 13m 23s & 41m 36s & 37m 44s & 9s  & 11s & 6h 31m 37s  & 2m 52s  & 2m 14s  & 33s\\
facebook  & 6m 34s  & 4m 29s  & 3m 56s  & 4h 45m 16s  & 1h 16m 44s  & 37s & 34s & 1h 13m 28s  & 5m 46s  & 16m 1s  & 30s\\
fashion & 38m5s & 18m 36s & 16m 51s & 43h 41m 5s  & 1h 55m 18s  & 37s & 10m 25s & 51m 23s & 6m 9s & 13m 5s  & 3m 53s\\
  \bottomrule
\end{tabular}
}
}
\end{table*}

\begin{table*}[!t]
\centering
  \caption{\hlr{Inference time.}}
  \label{tab:inference_time}
\resizebox{0.85\linewidth}{!}{
\hlr{
  \begin{tabular}{c|ccccc|cccccc}
    \toprule
Dataset & RRL & PLNN  & BNN & SAINT & FT  & CART  & LR  & SVM & RF  & XGBoost & LightGBM\\
\midrule
adni  & 0.770 & 0.858 & 0.703 & 7.820 & 7.890 & 0.271 & 0.119 & 1.069 & 6.619 & 0.362 & 0.445\\
adult & 0.826 & 0.462 & 0.341 & 0.210 & 0.340 & 0.017 & 0.106 & 56.279  & 3.630 & 0.339 & 0.241\\
bank-marketing  & 0.887 & 0.579 & 0.358 & 0.150 & 0.250 & 0.008 & 0.089 & 0.039 & 2.977 & 0.299 & 0.160\\
banknote  & 0.745 & 0.534 & 0.449 & 2.200 & 1.940 & 0.751 & 0.017 & 0.113 & 7.211 & 0.187 & 0.202\\
\midrule
activity  & 0.719 & 0.538 & 0.387 & 1.900 & 1.190 & 0.058 & 0.066 & 0.577 & 3.422 & 0.578 & 0.551\\
dota2 & 0.514 & 0.502 & 0.390 & 0.150 & 0.070 & 0.043 & 0.013 & 990.330 & 8.724 & 0.430 & 0.160\\
facebook  & 0.645 & 0.436 & 0.438 & 2.720 & 1.550 & 1.387 & 0.422 & 2.358 & 19.119  & 6.052 & 1.070\\
fashion & 0.879 & 0.593 & 0.373 & 2.370 & 0.220 & 0.124 & 0.086 & 592.065 & 6.925 & 1.371 & 1.396\\
  \bottomrule
\end{tabular}
}
}
\end{table*}

\section{Model Complexity}
\label{appendix:model_complexity}
\begin{figure}[t!]
    \centering
    \includegraphics[width=0.47\textwidth]{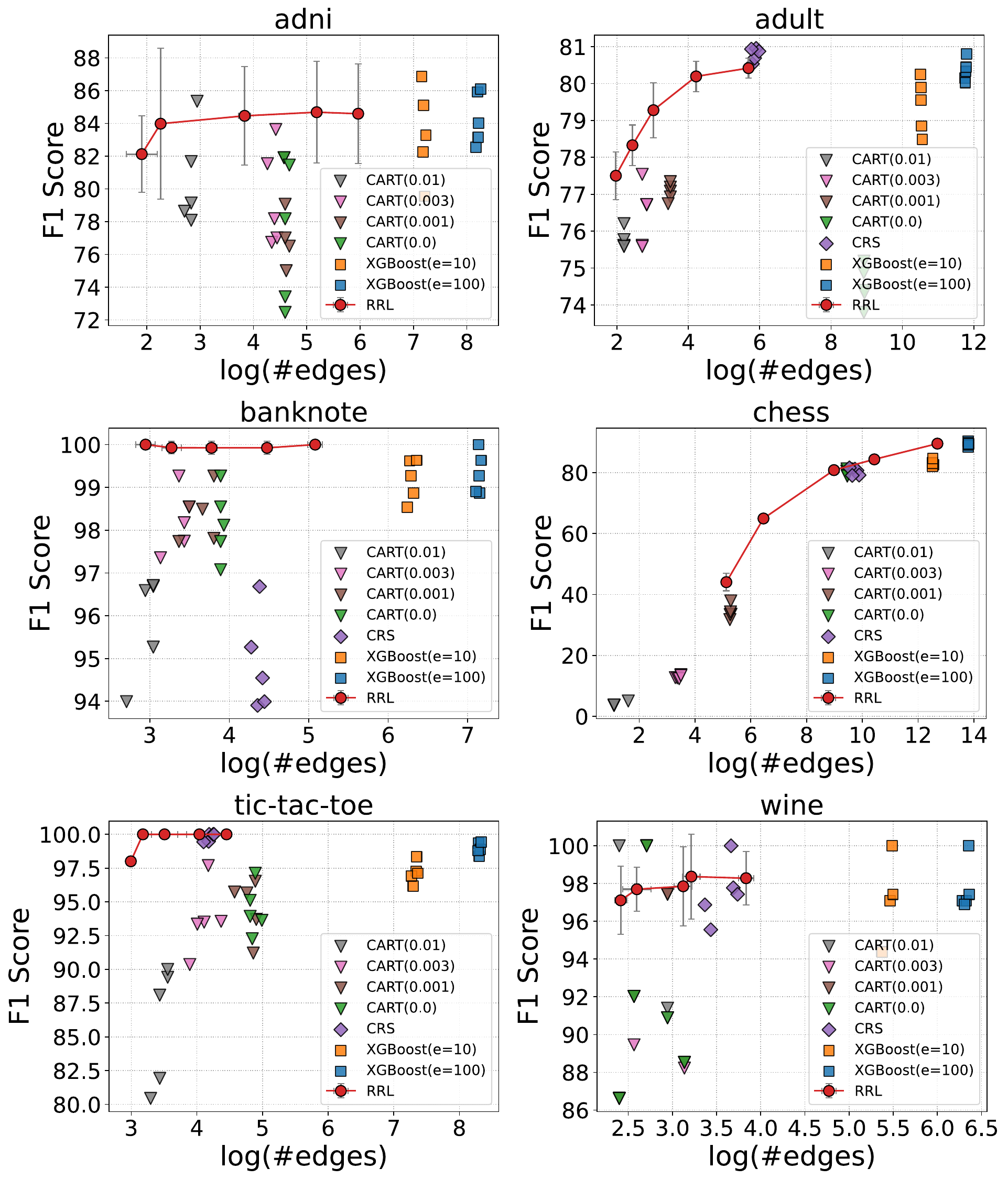}
    \caption{Scatter plot of F1 score against log(\#edges) for RRL and baselines on six datasets. F1 score and log(\#edges) are used to evaluate the classification performance and the model complexity, respectively.}
    \label{fig:supplementary_model_complexity}
\end{figure}

Figure \ref{fig:supplementary_model_complexity} shows the scatter plots of F1 score against log(\#edges) for rule-based models trained on the other six data sets. For RRL, the legend markers and error bars indicate means and standard deviations, respectively, of F1 score and log(\#edges) across cross-validation folds.
For baseline models, each point represents an evaluation of one model, on one fold, with one parameter setting.
The value in CART($\cdot$), e.g., CART(0.03), denotes the complexity parameter used for Minimal Cost-Complexity Pruning \cite{breiman2017classification}, and a higher value corresponds to a simpler tree. We also show the results of XGBoost with 10 and 100 estimators.
On these six data sets, we can still observe that if we connect the results of RRL, it will become a boundary that separates the upper left corner from other models. In other words, if RRL has a close model complexity with one baseline, then the F1 score of RRL will be higher, or if the F1 score of RRL is close to one baseline, then its model complexity will be lower. It indicates that RRL can make better use of rules than rule-based models using heuristic and ensemble methods in most cases.
\makeatletter
\setlength{\@fptop}{0pt}
\makeatother

\begin{figure}[t!]
    \centering
    \includegraphics[width=0.47\textwidth]{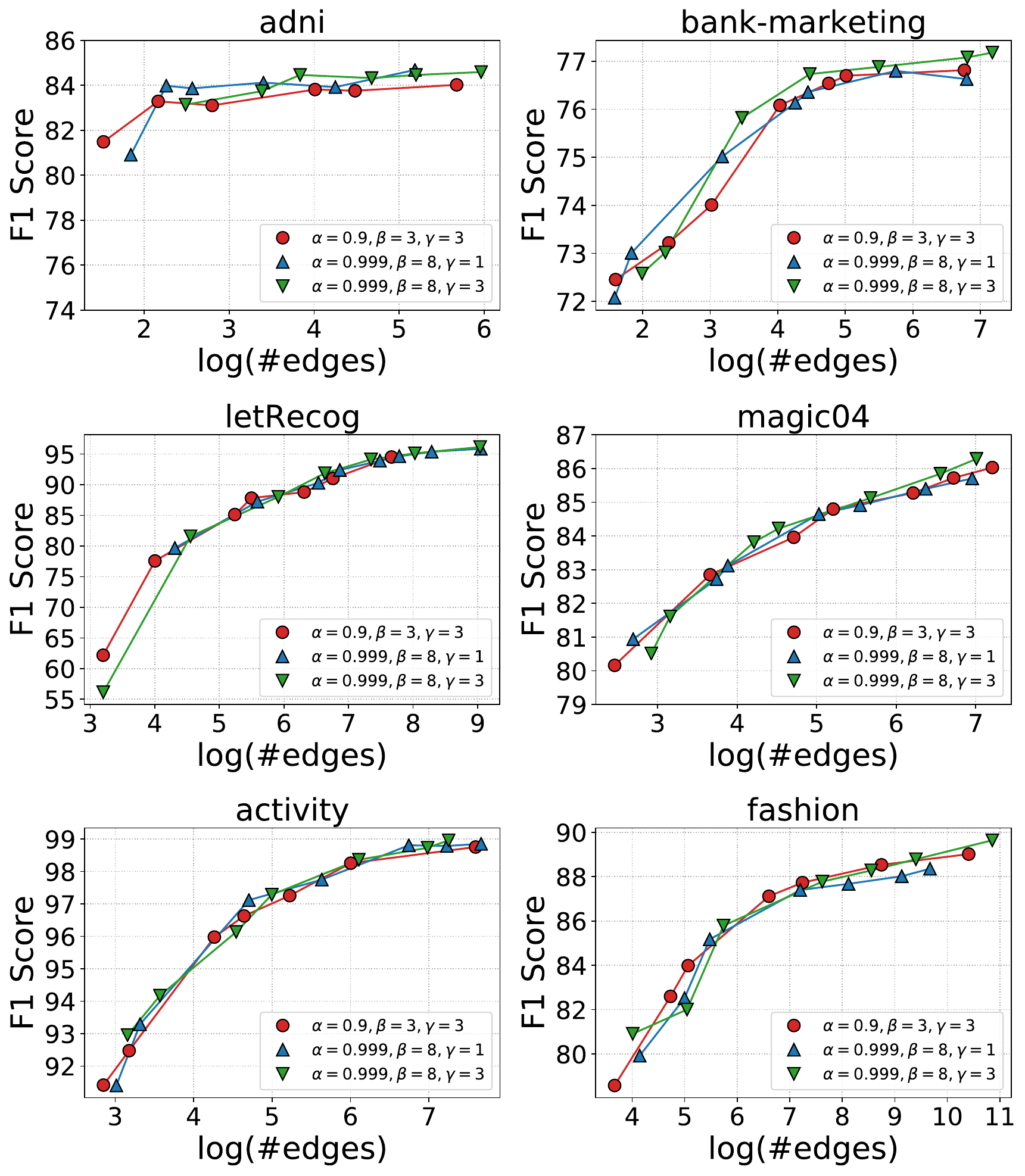}
    \caption{\hlr{F1 score and log(\#edges) of RRL using novel logical activation functions with different $(\alpha, \beta, \gamma)$ on six datasets. F1 score and log(\#edges) are used to evaluate the classification performance and the model complexity, respectively.}}
    \label{fig:alpha_beta_gamma_sensity}
\end{figure}

\hlr{
\section{Hyper-parameters $\alpha$, $\beta$, and $\gamma$}
\label{appendix:alpha_beta_gamma}
To further understand the behavior of the novel logical activation functions, we do sensitivity studies of the three hyper-parameters $\alpha$, $\beta$, and $\gamma$. As we mentioned in Section \ref{sec:NLAF}, these hyper-parameters have their own functions, and there are also requirements for their values. For example, $\alpha$ is used to change the value range and avoid the division by zero error, so 0.999 is better than 0.1 for $\alpha$. Therefore, we only test the values that meet the requirements in this part.
Take $(\alpha, \beta, \gamma)\in \{(0.999, 8, 1), (0.999, 8, 3), (0.9, 3, 3)\}$ as examples. The F1 score and model complexity of RRL using different $(\alpha, \beta, \gamma)$ on six datasets are shown in Figure \ref{fig:alpha_beta_gamma_sensity}.
We can see that the performance of RRL is not very sensitive if the values of $\alpha$, $\beta$, and $\gamma$ satisfy the requirements in most cases. Although the differences are not significant, for some datasets, a specific hyper-parameter selection could be more suitable for a specific model complexity range. Take the dataset \textit{adni} as an example. $(0.9, 3, 3)$ is more suitable for models with low complexity, i.e., using $(0.9, 3, 3)$ to set $(\alpha, \beta, \gamma)$ could lead to a higher F1 score when the model complexity is asked to be low. In addition, $(0.999, 8, 1)$ is more suitable for models with middle complexity, while $(0.999, 8, 3)$ is more suitable for models with high complexity. Other datasets also show the same phenomenon, more or less.
}

\begin{figure}[t!]
    \centering
    \includegraphics[width=0.47\textwidth]{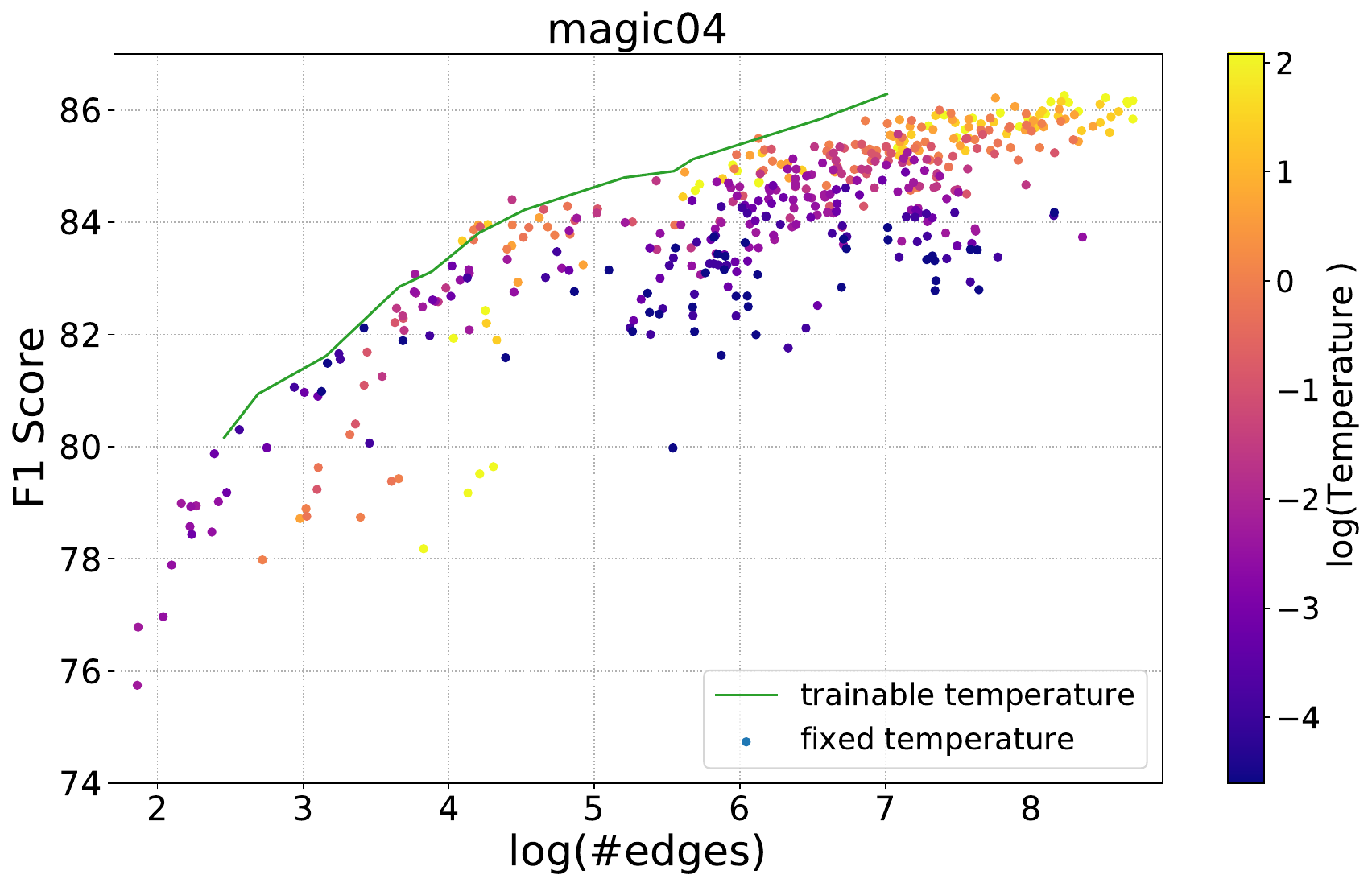}
    \caption{\hlr{Scatter plot of F1 score against log(\#edges) for RRL trained with fixed or trainable temperature. For RRL with a fixed temperature, one scatter represents an evaluation of one RRL with one parameter set, and the corresponding temperatures are depicted by color. F1 score and log(\#edges) are used to evaluate the classification performance and the model complexity, respectively.}}
    \label{fig:trainable_temp}
\end{figure}

\hlr{
\section{Temperature}
\label{appendix:temperature}
To show how a trainable temperature of the softmax in the loss function could benefit RRL, we compare RRL using a trainable temperature with RRL using fixed temperatures. On the \textit{magic04} dataset, we train RRL with the same structure but different learning rates, coefficients of the L2 regularization term, and $(\alpha, \beta, \gamma)$. The results are shown in Figure \ref{fig:trainable_temp}. For RRL with fixed temperature, one point represents an evaluation of one RRL with one parameter set, and the corresponding temperatures are depicted by color. For RRL with trainable temperature, we draw a green line to represent the performance. 
}

\hlr{
In Figure \ref{fig:trainable_temp}, we can first observe that, for RRL with fixed temperature, high temperature is suitable for RRL with high complexity, while low temperature is suitable for RRL with low complexity.
For example, when $\log(\#edge)>7$, RRL with $\log(Temperature)=2$ outperforms RRL with $\log(Temperature)=-2$. When $\log(\#edge)<5$, RRL with $\log(Temperature)=-2$ outperforms RRL with $\log(Temperature)=2$.
However, since we do not know the model complexity and its adaptable temperature in advance, if a fixed temperature is used, it can only be tuned as a hyperparameter, which dramatically increases the difficulty of training.
}

\hlr{
On the contrary, the trainable temperature can automatically adjust itself to adapt to different model complexities.
In Figure \ref{fig:trainable_temp}, we can see that the green line achieves the optimal (or better) results compared with those of fixed temperatures. Therefore, using trainable temperature can help RRL achieve a better trade-off between classification performance and model complexity.
Furthermore, the trainable temperature avoids tuning the hyperparameter temperature, simplifying RRL training.
}

\hlr{
\section{Case Study}
\label{appendix:case_study_fashion}
To show what the learned RRL looks like on a large dataset, we do case studies on the \textit{fashion} dataset. The \textit{fashion} dataset is used for an image classification task. 
Although RRL is not designed for image classification tasks, due to its high scalability, it can still provide intuition by visualizations and help us understand the decision mode. 
For each class, we combine the first ten rules, ordered by linear layer weights, for feature (pixel) visualization. In Figure \ref{fig:fashion}, a black/white pixel indicates the combined rule asks for a color close to black/white here in the original input image, and the grey pixel means no requirement in the rule. According to these figures, we can see how RRL classifies the images, e.g., distinguishing T-shirts from Pullovers by sleeves and distinguishing sneakers from ankle boots by heel.
\begin{figure}[t!]
    \centering
    \includegraphics[width=0.47\textwidth]{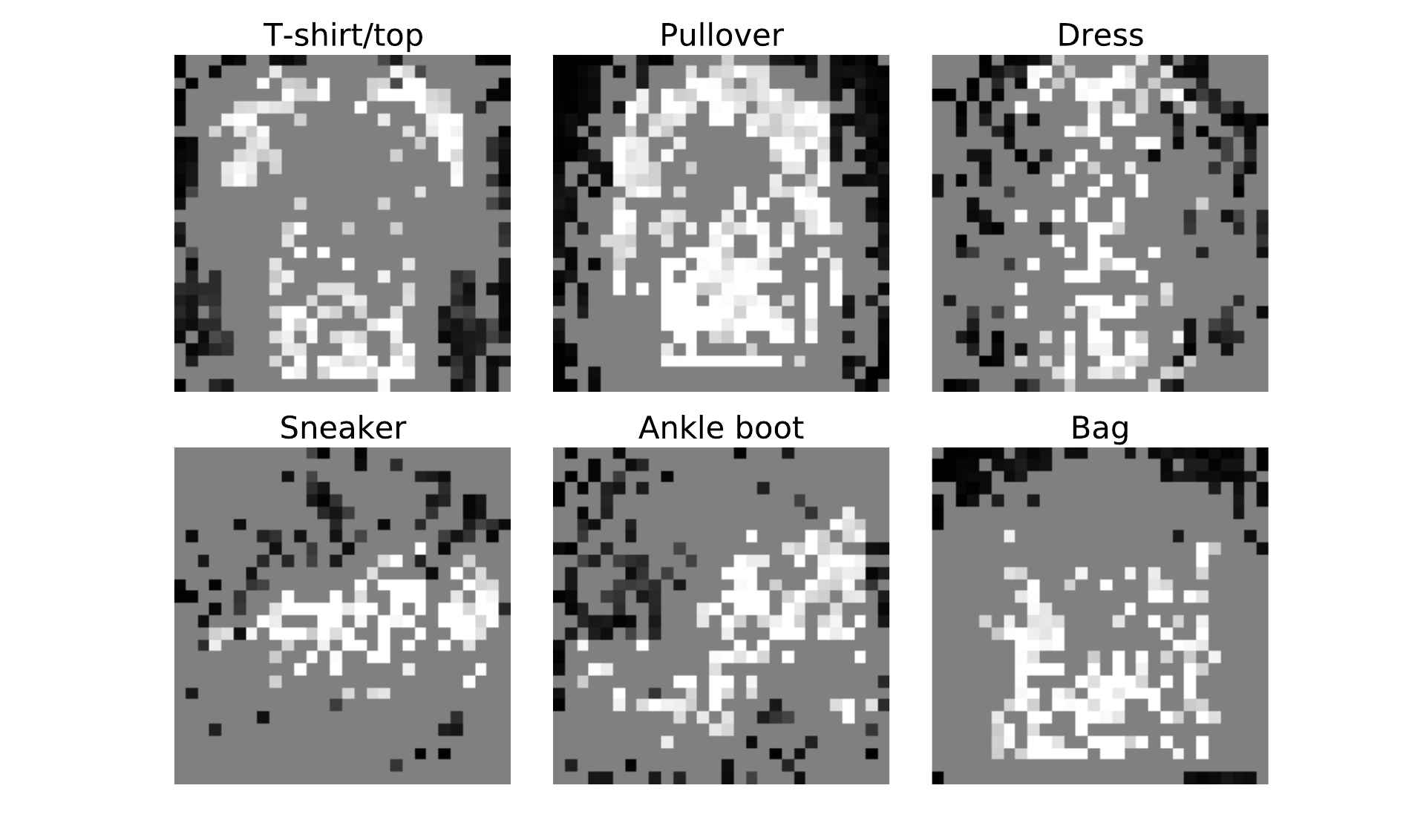}
    \caption{\hlr{Decision mode for the \textit{fashion} data set summarized from rules of RRL.}}
    \label{fig:fashion}
\end{figure}
}

\end{document}

%% file: math_commands.tex
\usepackage{amsmath,amsfonts,bm}

\def\eqref#1{equation~\ref{#1}}

\def\1{\bm{1}}

\def\va{{\bm{a}}}

\def\vc{{\bm{c}}}

\def\vn{{\bm{n}}}

\DeclareMathAlphabet{\mathsfit}{\encodingdefault}{\sfdefault}{m}{sl}
\SetMathAlphabet{\mathsfit}{bold}{\encodingdefault}{\sfdefault}{bx}{n}

\def\sD{{\mathbb{D}}}

\newcommand{\E}{\mathbb{E}}

\newcommand{\R}{\mathbb{R}}



%% file: bare_jrnl.bbl
\begin{thebibliography}{10}
\providecommand{\url}[1]{#1}
\csname url@samestyle\endcsname
\providecommand{\newblock}{\relax}
\providecommand{\bibinfo}[2]{#2}
\providecommand{\BIBentrySTDinterwordspacing}{\spaceskip=0pt\relax}
\providecommand{\BIBentryALTinterwordstretchfactor}{4}
\providecommand{\BIBentryALTinterwordspacing}{\spaceskip=\fontdimen2\font plus
\BIBentryALTinterwordstretchfactor\fontdimen3\font minus \fontdimen4\font\relax}
\providecommand{\BIBforeignlanguage}[2]{{%
\expandafter\ifx\csname l@#1\endcsname\relax
\typeout{** WARNING: IEEEtran.bst: No hyphenation pattern has been}%
\typeout{** loaded for the language `#1'. Using the pattern for}%
\typeout{** the default language instead.}%
\else
\language=\csname l@#1\endcsname
\fi
#2}}
\providecommand{\BIBdecl}{\relax}
\BIBdecl

\bibitem{goodfellow2016deep}
I.~Goodfellow, Y.~Bengio, A.~Courville, and Y.~Bengio, \emph{Deep learning}.\hskip 1em plus 0.5em minus 0.4em\relax MIT Press, 2016, vol.~1.

\bibitem{doshi2017towards}
F.~Doshi-Velez and B.~Kim, ``Towards a rigorous science of interpretable machine learning,'' \emph{arXiv preprint arXiv:1702.08608}, 2017.

\bibitem{molnar2019}
C.~Molnar, \emph{Interpretable Machine Learning}, 2019, \url{https://christophm.github.io/interpretable-ml-book/}.

\bibitem{lipton2016mythos}
Z.~C. Lipton, ``The mythos of model interpretability,'' \emph{Commun. {ACM}}, vol.~61, no.~10, pp. 36--43, 2018.

\bibitem{chu2018exact}
L.~Chu, X.~Hu, J.~Hu, L.~Wang, and J.~Pei, ``Exact and consistent interpretation for piecewise linear neural networks: A closed form solution,'' in \emph{SIGKDD}, 2018, pp. 1244--1253.

\bibitem{murdoch2019interpretable}
W.~J. Murdoch, C.~Singh, K.~Kumbier, R.~Abbasi-Asl, and B.~Yu, ``Interpretable machine learning: definitions, methods, and applications,'' \emph{PNAS}, vol. 116, no.~44, pp. 22\,071--22\,080, 2019.

\bibitem{ke2017lightgbm}
G.~Ke, Q.~Meng, T.~Finley, T.~Wang, W.~Chen, W.~Ma, Q.~Ye, and T.-Y. Liu, ``Lightgbm: A highly efficient gradient boosting decision tree,'' in \emph{NeurIPS}, 2017, pp. 3146--3154.

\bibitem{breiman2001random}
L.~Breiman, ``Random forests,'' \emph{Machine learning}, vol.~45, no.~1, pp. 5--32, 2001.

\bibitem{irsoy2012soft}
O.~Irsoy, O.~T. Y{\i}ld{\i}z, and E.~Alpayd{\i}n, ``Soft decision trees,'' in \emph{ICPR}, 2012, pp. 1819--1822.

\bibitem{letham2015interpretable}
B.~Letham, C.~Rudin, T.~H. McCormick, D.~Madigan \emph{et~al.}, ``Interpretable classifiers using rules and bayesian analysis: Building a better stroke prediction model,'' \emph{The Annals of Applied Statistics}, vol.~9, no.~3, pp. 1350--1371, 2015.

\bibitem{wang2017bayesian}
T.~Wang, C.~Rudin, F.~Doshi-Velez, Y.~Liu, E.~Klampfl, and P.~MacNeille, ``A bayesian framework for learning rule sets for interpretable classification,'' \emph{JMLR}, vol.~18, no.~1, pp. 2357--2393, 2017.

\bibitem{yang2017scalable}
H.~Yang, C.~Rudin, and M.~Seltzer, ``Scalable bayesian rule lists,'' in \emph{ICML}, 2017, pp. 3921--3930.

\bibitem{frosst2017distilling}
N.~Frosst and G.~Hinton, ``Distilling a neural network into a soft decision tree,'' \emph{arXiv preprint arXiv:1711.09784}, 2017.

\bibitem{ribeiro2016should}
M.~T. Ribeiro, S.~Singh, and C.~Guestrin, ``Why should i trust you?: Explaining the predictions of any classifier,'' in \emph{SIGKDD}, 2016, pp. 1135--1144.

\bibitem{wang2020transparent}
Z.~Wang, W.~Zhang, N.~Liu, and J.~Wang, ``Transparent classification with multilayer logical perceptrons and random binarization,'' in \emph{AAAI}, 2020, pp. 6331--6339.

\bibitem{cour2015bconnect}
M.~Courbariaux, Y.~Bengio, and J.-P. David, ``Binaryconnect: Training deep neural networks with binary weights during propagations,'' in \emph{NeurIPS}, 2015, pp. 3123--3131.

\bibitem{Quinlan:1993:CPM:583200}
J.~R. Quinlan, \emph{C4.5: Programs for Machine Learning}.\hskip 1em plus 0.5em minus 0.4em\relax San Francisco, CA, USA: Morgan Kaufmann Publishers Inc., 1993.

\bibitem{breiman2017classification}
L.~Breiman, \emph{Classification and regression trees}.\hskip 1em plus 0.5em minus 0.4em\relax Routledge, 2017.

\bibitem{cohen1995fast}
W.~W. Cohen, ``Fast effective rule induction,'' in \emph{MLP}.\hskip 1em plus 0.5em minus 0.4em\relax Elsevier, 1995, pp. 115--123.

\bibitem{wei2019generalized}
D.~Wei, S.~Dash, T.~Gao, and O.~Gunluk, ``Generalized linear rule models,'' in \emph{ICML}.\hskip 1em plus 0.5em minus 0.4em\relax PMLR, 2019, pp. 6687--6696.

\bibitem{angelino2017learning}
E.~Angelino, N.~Larus-Stone, D.~Alabi, M.~Seltzer, and C.~Rudin, ``Learning certifiably optimal rule lists for categorical data,'' \emph{JMLR}, vol.~18, no.~1, pp. 8753--8830, 2017.

\bibitem{lin2020generalized}
J.~Lin, C.~Zhong, D.~Hu, C.~Rudin, and M.~Seltzer, ``Generalized and scalable optimal sparse decision trees,'' in \emph{ICML}.\hskip 1em plus 0.5em minus 0.4em\relax PMLR, 2020, pp. 6150--6160.

\bibitem{lakkaraju2016interpretable}
H.~Lakkaraju, S.~H. Bach, and J.~Leskovec, ``Interpretable decision sets: A joint framework for description and prediction,'' in \emph{SIGKDD}, 2016, pp. 1675--1684.

\bibitem{chen2016xgboost}
T.~Chen and C.~Guestrin, ``Xgboost: A scalable tree boosting system,'' in \emph{SIGKDD}, 2016, pp. 785--794.

\bibitem{hara2016making}
S.~Hara and K.~Hayashi, ``Making tree ensembles interpretable: A bayesian model selection approach,'' in \emph{AISTATS}, 2018, pp. 77--85.

\bibitem{ishibuchi2005rule}
H.~Ishibuchi and T.~Yamamoto, ``Rule weight specification in fuzzy rule-based classification systems,'' \emph{TFS}, vol.~13, no.~4, pp. 428--435, 2005.

\bibitem{yang2018deep}
Y.~Yang, I.~G. Morillo, and T.~M. Hospedales, ``Deep neural decision trees,'' \emph{arXiv preprint arXiv:1806.06988}, 2018.

\bibitem{glanois2022neuro}
C.~Glanois, Z.~Jiang, X.~Feng, P.~Weng, M.~Zimmer, D.~Li, W.~Liu, and J.~Hao, ``Neuro-symbolic hierarchical rule induction,'' in \emph{ICML}, 2022, pp. 7583--7615.

\bibitem{cheng2022rlogic}
K.~Cheng, J.~Liu, W.~Wang, and Y.~Sun, ``Rlogic: Recursive logical rule learning from knowledge graphs,'' in \emph{SIGKDD}, 2022, pp. 179--189.

\bibitem{zimmer2021differentiable}
M.~Zimmer, X.~Feng, C.~Glanois, Z.~JIANG, J.~Zhang, P.~Weng, D.~Li, J.~HAO, and W.~Liu, ``Differentiable logic machines,'' \emph{TMLR}, 2023.

\bibitem{chaudhury-etal-2023-learning}
S.~Chaudhury, S.~Swaminathan, D.~Kimura, P.~Sen, K.~Murugesan, R.~Uceda-Sosa, M.~Tatsubori, A.~Fokoue, P.~Kapanipathi, A.~Munawar, and A.~Gray, ``Learning symbolic rules over {A}bstract {M}eaning {R}epresentations for textual reinforcement learning,'' in \emph{ACL}, 2023, pp. 6764--6776.

\bibitem{duan2022deeplogic}
X.~Duan, X.~Wang, P.~Zhao, G.~Shen, and W.~Zhu, ``Deeplogic: Joint learning of neural perception and logical reasoning,'' \emph{TPAMI}, vol.~45, no.~4, pp. 4321--4334, 2023.

\bibitem{zhou2021abductive}
Z.-H. Zhou and Y.-X. Huang, ``Abductive learning,'' in \emph{Neuro-Symbolic Artificial Intelligence: The State of the Art}.\hskip 1em plus 0.5em minus 0.4em\relax IOS Press, 2021, pp. 353--369.

\bibitem{dai2019bridging}
W.~Dai, Q.~Xu, Y.~Yu, and Z.~Zhou, ``Bridging machine learning and logical reasoning by abductive learning,'' pp. 2811--2822, 2019.

\bibitem{zhang2020mining}
Q.~Zhang, J.~Ren, G.~Huang, R.~Cao, Y.~N. Wu, and S.-C. Zhu, ``Mining interpretable aog representations from convolutional networks via active question answering,'' \emph{TPAMI}, vol.~43, no.~11, pp. 3949--3963, 2020.

\bibitem{Liu23fire}
B.~Liu and R.~Mazumder, ``Fire: An optimization approach for fast interpretable rule extraction,'' in \emph{SIGKDD}, 2023, p. 1396–1405.

\bibitem{courbariaux2016binarized}
I.~Hubara, M.~Courbariaux, D.~Soudry, R.~El{-}Yaniv, and Y.~Bengio, ``Binarized neural networks,'' in \emph{NeurIPS}, 2016, pp. 4107--4115.

\bibitem{bai2018proxquant}
Y.~Bai, Y.-X. Wang, and E.~Liberty, ``Proxquant: Quantized neural networks via proximal operators,'' \emph{arXiv preprint arXiv:1810.00861}, 2018.

\bibitem{jang2016categorical}
E.~Jang, S.~Gu, and B.~Poole, ``Categorical reparameterization with gumbel-softmax,'' 2017.

\bibitem{payani2019learning}
A.~Payani and F.~Fekri, ``Learning algorithms via neural logic networks,'' \emph{arXiv preprint arXiv:1904.01554}, 2019.

\bibitem{hajek2013metamathematics}
P.~H{\'a}jek, \emph{Metamathematics of fuzzy logic}.\hskip 1em plus 0.5em minus 0.4em\relax Springer Science \& Business Media, 2013, vol.~4.

\bibitem{Learning279181}
Y.~Bengio, P.~Simard, and P.~Frasconi, ``Learning long-term dependencies with gradient descent is difficult,'' \emph{TNNS}, vol.~5, no.~2, pp. 157--166, 1994.

\bibitem{abadi2016tensorflow}
M.~Abadi, P.~Barham, J.~Chen, Z.~Chen, A.~Davis, J.~Dean, M.~Devin, S.~Ghemawat, G.~Irving, M.~Isard \emph{et~al.}, ``$\{$TensorFlow$\}$: a system for $\{$Large-Scale$\}$ machine learning,'' in \emph{OSDI}, 2016, pp. 265--283.

\bibitem{paszke2019pytorch}
A.~Paszke, S.~Gross, F.~Massa, A.~Lerer, J.~Bradbury, G.~Chanan, T.~Killeen, Z.~Lin, N.~Gimelshein, L.~Antiga \emph{et~al.}, ``Pytorch: An imperative style, high-performance deep learning library,'' in \emph{NeurIPS}, 2019, pp. 8026--8037.

\bibitem{cook2012cuda}
S.~Cook, \emph{CUDA programming: a developer's guide to parallel computing with GPUs}.\hskip 1em plus 0.5em minus 0.4em\relax Newnes, 2012.

\bibitem{qin2020binary}
H.~Qin, R.~Gong, X.~Liu, X.~Bai, J.~Song, and N.~Sebe, ``Binary neural networks: A survey,'' \emph{Pattern Recognition}, p. 107281, 2020.

\bibitem{hinton2015distilling}
G.~Hinton, O.~Vinyals, J.~Dean \emph{et~al.}, ``Distilling the knowledge in a neural network,'' \emph{arXiv preprint arXiv:1503.02531}, vol.~2, no.~7, 2015.

\bibitem{Dua:2019}
\BIBentryALTinterwordspacing
D.~Dua and C.~Graff, ``{UCI} machine learning repository,'' 2017. [Online]. Available: \url{http://archive.ics.uci.edu/ml}
\BIBentrySTDinterwordspacing

\bibitem{xiao2017:online}
H.~Xiao, K.~Rasul, and R.~Vollgraf, ``Fashion-mnist: a novel image dataset for benchmarking machine learning algorithms,'' 2017.

\bibitem{anguita2013public}
D.~Anguita, A.~Ghio, L.~Oneto, X.~Parra, J.~L. Reyes-Ortiz \emph{et~al.}, ``A public domain dataset for human activity recognition using smartphones.'' in \emph{Esann}, vol.~3, 2013, p.~3.

\bibitem{rozemberczki2019multiscale}
B.~Rozemberczki, C.~Allen, and R.~Sarkar, ``{Multi-Scale attributed node embedding},'' \emph{Journal of Complex Networks}, vol.~9, no.~2, p. cnab014, 05 2021.

\bibitem{petersen2010alzheimer}
R.~C. Petersen, P.~Aisen, L.~A. Beckett, M.~Donohue, A.~Gamst, D.~J. Harvey, C.~Jack, W.~Jagust, L.~Shaw, A.~Toga \emph{et~al.}, ``Alzheimer's disease neuroimaging initiative (adni): clinical characterization,'' \emph{Neurology}, vol.~74, no.~3, pp. 201--209, 2010.

\bibitem{wang2022learning}
Z.~Wang, J.~Wang, N.~Liu, C.~Liu, X.~Li, L.~Dong, R.~Zhang, C.~Mao, Z.~Duan, W.~Zhang \emph{et~al.}, ``Learning cognitive-test-based interpretable rules for prediction and early diagnosis of dementia using neural networks,'' \emph{Journal of Alzheimer's Disease}, vol.~90, no.~2, pp. 609--624.

\bibitem{demvsar2006statistical}
J.~Dem{\v{s}}ar, ``Statistical comparisons of classifiers over multiple data sets,'' \emph{JMLR}, vol.~7, no. Jan, pp. 1--30, 2006.

\bibitem{kingma2014adam}
D.~P. Kingma and J.~Ba, ``Adam: {A} method for stochastic optimization,'' 2015.

\bibitem{kleinbaum2002logistic}
D.~G. Kleinbaum, K.~Dietz, M.~Gail, M.~Klein, and M.~Klein, \emph{Logistic regression}.\hskip 1em plus 0.5em minus 0.4em\relax Springer, 2002.

\bibitem{GorishniyRKB21}
Y.~Gorishniy, I.~Rubachev, V.~Khrulkov, and A.~Babenko, ``Revisiting deep learning models for tabular data,'' in \emph{NeurIPS}, 2021, pp. 18\,932--18\,943.

\bibitem{somepalli2021saint}
G.~Somepalli, M.~Goldblum, A.~Schwarzschild, C.~B. Bruss, and T.~Goldstein, ``Saint: Improved neural networks for tabular data via row attention and contrastive pre-training,'' \emph{arXiv preprint arXiv:2106.01342}, 2021.

\bibitem{scholkopf2001learning}
B.~Scholkopf and A.~J. Smola, \emph{Learning with kernels: support vector machines, regularization, optimization, and beyond}.\hskip 1em plus 0.5em minus 0.4em\relax MIT press, 2001.

\bibitem{nair2010rectified}
V.~Nair and G.~E. Hinton, ``Rectified linear units improve restricted boltzmann machines,'' in \emph{ICML}, 2010, pp. 807--814.

\end{thebibliography}
